\documentclass[10pt,twocolumn,letterpaper]{article}

\usepackage[pagenumbers]{cvpr} 
\def\paperID{} 

\makeatletter
\@namedef{ver@everyshi.sty}{}
\makeatother

\usepackage{graphicx}
\usepackage{tikz}
\usepackage{comment} 
\usepackage{color}

\usepackage{booktabs}
\usepackage{times}
\usepackage{epsfig}
\usepackage{mathptmx}
\usepackage{graphicx}
\usepackage{amsmath}
\usepackage{amsthm}
\usepackage{amssymb}

\usepackage{wrapfig}

\usepackage[pagebackref,breaklinks,colorlinks,citecolor=cvprblue]{hyperref} 

\def\confName{CVPR}
\def\confYear{2024}

\usepackage{xcolor}
\definecolor{cvprblue}{rgb}{0.21,0.49,0.74}
\definecolor{fb_blue}{RGB}{66 103 178}
\definecolor{fb_grey}{RGB}{137 143 156}
\definecolor{ins_yellow}{RGB}{255,220,128}
\definecolor{ins_orange}{RGB}{247,119,55}

\xdefinecolor{navyblue}{RGB}{19 41 75}
\xdefinecolor{carolinablue}{RGB}{123 175 212}
\xdefinecolor{cb}{RGB}{123 175 212}
\xdefinecolor{flatgrey}{RGB}{162 170 173}
\xdefinecolor{matcha}{RGB}{183 224 176}

\usepackage{tikz}
\tikzset{arrow/.style={-stealth, thick, draw=gray!80!black}}
\usetikzlibrary{decorations.pathreplacing,calligraphy,perspective,3d}
\usetikzlibrary{positioning}
\usetikzlibrary{shapes.geometric}   
\usepackage{rotating}               
\usepackage{changepage}
\usepackage{tikz-3dplot}
\usepackage{tikzscale}
\usetikzlibrary{plotmarks}
\usepackage{pgfplots}
\usetikzlibrary{tikzmark,calc} 
\usetikzlibrary{decorations.pathreplacing, decorations.markings}
\usepackage{movie15}
\usepackage{animate}
  
\usepackage[capbesideposition=right]{floatrow}

\newcommand{\dashedLayer}[6]{
			\def\a{#1} 
			\def\b{0.02}
			\def\c{#2} 
			\def\t{#3} 
			\def\d{#4} 

			\draw[line width=0.15mm, dash pattern=on 2.25pt off 1.8pt](\c+\t,0,\d) -- (\c+\t,\a,\d) -- (\t,\a,\d);                                                      
			\draw[line width=0.15mm, dash pattern=on 2.25pt off 1.8pt](\t,0,\a+\d) -- (\c+\t,0,\a+\d) node[midway,below] {#6} -- (\c+\t,\a,\a+\d) -- (\t,\a,\a+\d) -- (\t,0,\a+\d); 
			\draw[line width=0.15mm, dash pattern=on 2.25pt off 1.8pt](\c+\t,0,\d) -- (\c+\t,0,\a+\d);
			\draw[line width=0.15mm, dash pattern=on 2.25pt off 1.8pt](\c+\t,\a,\d) -- (\c+\t,\a,\a+\d);
			\draw[line width=0.15mm, dash pattern=on 2.25pt off 1.8pt](\t,\a,\d) -- (\t,\a,\a+\d);
			
			\draw[line width=0.15mm] (\c+\t,0,\d) -- (\c+\t,\a,\d);
			\draw[line width=0.15mm] (\c+\t,0,\d) -- (\c+\t,0,\a+\d);
			\draw[line width=0.15mm] (\c+\t,\a,\d) -- (\c+\t,\a,\a+\d);
			\draw[line width=0.15mm] (\c+\t,0,\a+\d) -- (\c+\t,\a,\a+\d);
			
			\filldraw[#5] (\t+\b,\b,\a+\d) -- (\c+\t-\b,\b,\a+\d) -- (\c+\t-\b,\a-\b,\a+\d) -- (\t+\b,\a-\b,\a+\d) -- (\t+\b,\b,\a+\d); 
			\filldraw[#5] (\t+\b,\a,\a-\b+\d) -- (\c+\t-\b,\a,\a-\b+\d) -- (\c+\t-\b,\a,\b+\d) -- (\t+\b,\a,\b+\d);
			\ifthenelse {\equal{#5} {}}
			{} 
			{\filldraw[#5] (\c+\t,\b,\a-\b+\d) -- (\c+\t,\b,\b+\d) -- (\c+\t,\a-\b,\b+\d) -- (\c+\t,\a-\b,\a-\b+\d);} 
		}

\newcommand{\networkLayer}[6]{
			\def\a{#1} 
			\def\b{0.02}
			\def\c{#2} 
			\def\t{#3} 
			\def\d{#4} 

			\draw[line width=0.15mm](\c+\t,0,\d) -- (\c+\t,\a,\d) -- (\t,\a,\d);                                                      
			\draw[line width=0.15mm](\t,0,\a+\d) -- (\c+\t,0,\a+\d) node[midway,below] {#6} -- (\c+\t,\a,\a+\d) -- (\t,\a,\a+\d) -- (\t,0,\a+\d); 
			\draw[line width=0.15mm](\c+\t,0,\d) -- (\c+\t,0,\a+\d);
			\draw[line width=0.15mm](\c+\t,\a,\d) -- (\c+\t,\a,\a+\d);
			\draw[line width=0.15mm](\t,\a,\d) -- (\t,\a,\a+\d);

			\filldraw[#5] (\t+\b,\b,\a+\d) -- (\c+\t-\b,\b,\a+\d) -- (\c+\t-\b,\a-\b,\a+\d) -- (\t+\b,\a-\b,\a+\d) -- (\t+\b,\b,\a+\d); 
			\filldraw[#5] (\t+\b,\a,\a-\b+\d) -- (\c+\t-\b,\a,\a-\b+\d) -- (\c+\t-\b,\a,\b+\d) -- (\t+\b,\a,\b+\d);
			\ifthenelse {\equal{#5} {}}
			{} 
			{\filldraw[#5] (\c+\t,\b,\a-\b+\d) -- (\c+\t,\b,\b+\d) -- (\c+\t,\a-\b,\b+\d) -- (\c+\t,\a-\b,\a-\b+\d);} 
		}
		
\usepackage[linesnumbered,ruled,vlined]{algorithm2e}

\SetCommentSty{mycommfont}
\SetKwInput{Input}{Input}                
\SetKwInput{Output}{Output}         
\SetKwInput{Dataset}{Dataset}     
\SetKwInput{Settings}{Settings}      
\SetKwInput{Initialization}{Initialization}      
\SetAlgorithmName{Alg.}{Alg.}{List of Algorithms} 
\makeatletter
\newcommand{\algorithmfootnote}[2][\footnotesize]{%
  \let\old@algocf@finish\@algocf@finish
  \def\@algocf@finish{\old@algocf@finish
    \leavevmode\rlap{\begin{minipage}{\linewidth}
    #1#2
    \end{minipage}}%
  }%
}
\makeatother

\usepackage{floatrow}

\usepackage{caption} 
\usepackage{subcaption}
\usepackage{makecell}
\usepackage{booktabs}
\usepackage{multirow}

\newcolumntype{P}[1]{>{\centering\arraybackslash}p{#1}}

\usepackage{todonotes}

\usepackage{hyperref} 
  
\newtheoremstyle{bfnote}%
  {}{}
  {\itshape}{}
  {\bfseries}{.}
  { }{\thmname{#1}\thmnumber{ #2}\thmnote{ (#3)}}
\theoremstyle{bfnote}

\usepackage{url}

\definecolor{olive}{rgb}{0.42, 0.56, 0.14}

\pgfplotsset{compat=1.18}
\begin{document}

\title{Differential Motion Evolution for Fine-Grained Motion Deformation in Unsupervised Image Animation}

\author{Peirong Liu\textsuperscript{1} \quad Rui Wang\textsuperscript{2} \quad Xuefei Cao\textsuperscript{2} \quad Yipin Zhou\textsuperscript{2} \quad Ashish Shah\textsuperscript{2}  \quad Ser-Nam Lim\textsuperscript{2} \vspace{0.3cm} \\  
\textsuperscript{1}University of North Carolina at Chapel Hill \quad \textsuperscript{2}Meta AI \vspace{0.3cm} 
}
\maketitle

 

\begin{abstract}
\vspace{-0.1cm}
   
Image animation is the task of transferring the motion of a driving video to a given object in a source image. While great progress has recently been made in unsupervised motion transfer, requiring no labelled data or domain priors, many current unsupervised approaches still struggle to capture the motion deformations when large motion/view discrepancies occur between the source and driving domains. Under such condition, there is simply not enough information to capture the motion field properly. We introduce \texttt{DiME} (\textbf{Di}fferential \textbf{M}otion \textbf{E}volution), an end-to-end unsupervised motion transfer framework integrating differential refinement for motion estimation. Key findings are twofold: (1) by capturing the motion transfer with an ordinary differential equations (ODE), it helps to regularize the motion field, and (2) by utilizing the source image itself, we are able to inpaint occluded/missing regions arising from large motion changes. Additionally, we also propose a natural extension to the ODE idea, which is that \texttt{DiME} can easily leverage multiple different views of the source object whenever they are available by modeling an ODE per view. Extensive experiments across 9 benchmarks show \texttt{DiME} outperforms the state-of-the-arts by a significant margin and generalizes much better to unseen objects.
\end{abstract}

\vspace{-0.3cm}

\section{Introduction}
\label{sec: intro}  

\vspace{-0.15cm}



Motion transfer (or image animation) animates a source object according to the motion derived from a driving video, where the object identities in the source and driving images could be different. It has a wide range of applications including entertainment~\cite{Siarohin_2019_CVPR,Siarohin_2021_CVPR}, and video conferencing~\cite{Zakharov_2020_ECCV,Wang_2021_CVPR,Oquab_2020_arxiv}. Traditional motion transfer approaches typically require a large amount of labeled data, e.g., manually-selected landmarks, semantic segmentations, and parametric 3D models~\cite{Cao_2014_TG,Blanz_1999_ACCGIT,Thies_2016_CVPR,Zollhofer_2018_CGF,Qian_2019_ICCV,Zakharov_2019_CVPR}. Recent deep generative models such as GAN/VAE-based work avoids the need of object modeling but still depends on pre-trained object-specific representation models such as keypoints, poses and shapes extractors~\cite{Cao_2017_CVPR,Wang_2018_Neurips,Chan_2019_ICCV,Geng_2019_CVPR,Ren_2021_CVPR}. 

Most recently, a stream of unsupervised motion transfer approaches have been proposed to avoid relying on specific domain priors or significant labeling efforts, by leveraging the unsupervised key-points~\cite{Wiles_2018_ECCV,Siarohin_2019_CVPR,Siarohin_2019_NeurIPS,Wang_2021_CVPR} or regions~\cite{Siarohin_2019_CVPR} extractors to approximate the motion deformation fields between the source and driving domains. An encoder-decoder network is used to extract features from the source image, which are then warped to the driving domain according to the predicted deformation during generation. These unsupervised approaches could be trained end-to-end without additional domain priors, making it possible to animate a broader range of objects. 

However, without pre-specified models (e.g., 3D morphable models~\cite{Blanz_1999_ACCGIT}) or domain priors (e.g., extracted poses~\cite{Wang_2018_Neurips,Ren_2021_CVPR}) of the source object as additional inputs, the above unsupervised approaches still struggle 
when there is large motion changes between the source and the driving video (e.g., source is a left side view of a face whereas the driving video is a frontal view): 1) the key-point/region-based transformation only provides a \textit{coarse} estimation of the motion deformation, and is insufficient to capture more fine-grained motions; 2) without additional information, the missing regions under the driving view need to be inpainted by the image generator \textit{from scratch}, which could lead to visually unrealistic results (Fig.~\ref{fig: anim}, Fig.~\ref{fig: vox_mgif_gen}).

We propose \texttt{DiME} (\textbf{Di}fferential \textbf{M}otion \textbf{E}volution), an end-to-end unsupervised framework for motion transfer, switchable between single source (one-shot) and multiple sources (few-shot) setups during both training and inference. We highlight the capabilities of \texttt{DiME} as follows:

\vspace{-0.4cm}
\paragraph{Motion transfer by regularized optimization.}
We re-formulate the motion estimation task as a \textit{regularized optimization} problem, constrained by a system of regularized ODEs, where the number of ODEs is determined by the number of reference images of the source object, naturally resulting in a flexible one/few-shot setup.\footnote{In the one-shot setup, since no reference image is used, the ODE system reduces to a single ODE.} Each ODE parameterizes the evolution dynamics of the motion deformation from the corresponding source/reference image to the driving domain. The integration of the ODE system thus acts as a smooth \textit{refinement} of the estimated motion deformation fields~(Sec.~\ref{sec: refine}), while avoiding the expensive sparse-to-dense flow computations (e.g., Jacobians in~\cite{Siarohin_2019_CVPR,Wang_2021_CVPR}, singular value decompositions (SVD) in~\cite{Siarohin_2021_CVPR}).

\vspace{-0.4cm}
\paragraph{Source identity conditioned motion warping.} We further condition the motion deformation on the source identity, by extracting a ``self-flow'' field from the source itself, which imposes information from regions with similar appearance within the source object to help inpaint the occluded/missing regions under the driving domain~(Sec.~\ref{sec: appearance}).

\vspace{-0.4cm}
\paragraph{Flexible multi-view fusion during inference.}
Designed in an integrated ODE system, \texttt{DiME} is \textit{not} restricted by the number of source images to be used during inference time. Depending on specific application scenarios, one could either opt for the one-shot setup, or choose the few-shot setup to further boost the model's performance~(Sec.~\ref{sec: num_refs}). 
To the best of our knowledge, \texttt{DiME} is the \textit{first} few-shot motion transfer model that does \textit{not} rely on prior object representations, making it possible to be generalized to a broader range of objects (Sec.~\ref{sec: generality}, Fig.~\ref{fig: vox_mgif_gen}).

\vspace{0.15cm}
Extensive experiments across \emph{nine} datasets containing human faces, human bodies, robots, and cartoon animals consistently demonstrate that \texttt{DiME} outperforms the state-of-the-arts by a significant margin (Sec.~\ref{sec: comparisons}, Tab.~\ref{tab: all}). Ablation studies explore the effectiveness of \texttt{DiME}'s design in detail~(Sec.~\ref{sec: ablation}). Further, we test \texttt{DiME}'s performance on zero-shot motion transfer, i.e., transferring motion to novel objects that have not been seen during training, \texttt{DiME} generalises the best and performs the stablest especially when dealing with large motion changes~(Sec.~\ref{sec: ablation}, Fig.~\ref{fig: vox_mgif_gen}).

\vspace{-0.2cm}

\section{Related Work}
\label{sec: related}

\vspace{-0.1cm}

\paragraph{Image Animation}
\label{sec: rel_anim}
Image animation transfers the motion information from one driving video to a source image, where the identities in the source image and driving video are not necessarily the same. Traditional supervised approaches require pre-specified landmarks, segmentations or 3D models~\cite{Cao_2014_TG,Blanz_1999_ACCGIT,Thies_2016_CVPR,Zollhofer_2018_CGF,Qian_2019_ICCV,Zakharov_2019_CVPR,Geng_2019_CVPR}. As a result, these approaches usually need a large amount of labelled data, restricting the model to specific domains (e.g., faces, bodies, etc). Recently, several unsupervised approaches have been proposed to address the above challenges on the model flexibility~\cite{Siarohin_2019_CVPR,Siarohin_2021_CVPR,Siarohin_2019_NeurIPS,Wang_2021_CVPR,Wang_2019_Neurips,Zakharov_2020_ECCV,Wiles_2018_ECCV}. Monkey-Net~\cite{Siarohin_2019_CVPR} learns a set of key-points for sparse motions prediction in an unsupervised manner. The first order motion model (FOMM)~\cite{Siarohin_2019_NeurIPS} improves the motion estimation by resorting to local affine transformations. The articulated-animation model (AA)~\cite{Siarohin_2021_CVPR} is also designed to improve the motion estimation, but instead proposing to apply the principal component analysis (PCA) on the sparse motions' heatmaps. By leveraging unsupervised approaches, these models are able to scale image animation to a wider range of objects, e.g., faces, bodies, robots, and cartoon animals~\cite{Siarohin_2019_CVPR,Siarohin_2019_NeurIPS}. 


\vspace{-0.3cm}

\paragraph{Few-Shot Motion Retargeting}
\label{sec: rel_multi}
For better generalization to unseen domains and handling large motion changes, models have been proposed to take more than one image of the source object with arbitrary poses. This few-shot strategy has been recently used in face reenactment~\cite{Ha_2020_AAAI,Qian_2019_ICCV,Zakharov_2019_CVPR}, body retargeting~\cite{Lee_2019_ICLR,Wang_2019_Neurips,Ren_2021_CVPR,Loper_2015_SIGGRAPH,Li_2019_CVPR,Liu_2019_ICCV}, clothing transfer~\cite{Bhatnagar_2019_ICCV,Ma_2020_CVPR,Mir_2020_CVPR,Patel_2020_CVPR}, etc. However, all these works \textit{rely} on specific object representations such as 3D meshes, facial landmarks or body poses as conditional inputs, provided by pre-trained models. This limits the few-shot retargeting settings for a broader applications on objects whose representations are not well-explored. \texttt{DiME}, on the other hand, does \textit{not} need any additional pre-defined object representation. Given only one or a few images of the source object and a driving video, \texttt{DiME} integrates a system of ODEs forward to obtain a fine-grained motion deformation field from the source to the driving domain, in an entirely \textit{unsupervised} and \textit{end-to-end} fashion, making it possible for \texttt{DiME} to perform motion transfer on a much broader range of objects in addition to faces/bodies, such as cloths, robots, or even cartoon characters (Sec.~\ref{sec: datasets}).


\vspace{-0.3cm}

\paragraph{Neural-ODEs} 
\label{sec: rel_node}

Ordinary differential equations (ODEs) are widely used to represent the evolution of system dynamics. Models such as residual networks and recurrent neural networks, represent complex transformations by predicting a sequence of changes as the hidden states~\cite{Lu_2018_ICML,Ruthotto_2018_JMIV}. Chen et al. ~\cite{Chen_2018_Neurips} first proposed the neural-ODEs to parameterize the continuous dynamics of hidden units using neural networks. 
ANODE~\cite{Zhang_2019_Neurips} further addressed the instability issue of neural-ODEs by Discretize-Then-Optimize (DTO) differentiation. Going beyond the transport phenomena in physics and chemistry, neural-ODEs have been used in computer vision tasks, e.g., image segmentation~\cite{Rafael_2019_arxiv}, reconstruction~\cite{Ali_2019_arxiv}, super-resolution~\cite{Paoletti_2020_TGRS}, classification~\cite{He_2019_CVPR}, and medical imaging~\cite{Liu_2021_CVPR,Liu_2022_CVPR}, where neural-ODEs succeeded in capturing complex transformations via the continuous evolution of the system dynamics. In \texttt{DiME}, we formulate the few-shot motion estimation problem in an ODE system, where each ODE solves the coarse-to-fine refinement of the motion deformation fields between the source and driving domains~(Sec.~\ref{sec: refine}). This differential motion evolution strategy also naturally avoids the computationally expensive Jacobians or SVD operations~\cite{Siarohin_2019_NeurIPS,Siarohin_2021_CVPR,Wang_2021_CVPR} required for dense motion approximation.


\vspace{-0.3cm}

\paragraph{Regularized Optimal Mass Transport and Energy Minimization}
\label{sec: rel_omt}

Optimal mass transport (OMT)~\cite{Kantorovitch1958OnTT} treats the problem of optimally transporting mass distribution from one configuration to another via the minimization of a target (cost) function. 
Benamou and Brenier reformulated OMT in a computational fluid dynamics (CFD) framework~\cite{benamou2000omt}, where OMT was proven to be equal to a regularized energy minimization problem with a continuity constraint, e.g., a PDE system. Recently, OMT theory has received extensive research attention with applications in machine learning~\cite{Torres2021ASO}, image processing/registration~\cite{Feydy2017OptimalTF,Fitschen2016TransportBR,Haker2001MassPM}, network theory~\cite{Buttazzo2009AnOP}, biomedical science~\cite{Zhang2021ARO,Koundal2020OptimalMT}, and the Schr\"odinger bridge and entropic regularization~\cite{Chen2016OnTR,NIPS2013_af21d0c9}, etc. In \texttt{DiME}, we are interested in formulating the motion transfer task as a regularized minimization problem, yet in the context of motion transfer, the target is minimizing the motion differences between the source and driving images, with the associated continuity constraint representing the coarse-to-fine evolution of the estimated motion deformation (Sec.~\ref{sec: setup}).



\vspace{-0.cm}

\section{\texttt{DiME}: Differential Motion Evolution}
\label{sec: method}
\vspace{-0.1cm}

In Sec.~\ref{sec: setup}, we describe the problem setup where the motion transfer task is presented as a regularized optimization problem, and provide an overview on \texttt{DiME}'s workflow. In Sec.~\ref{sec: ode}, we detail our proposed coarse-to-fine motion evolution in an ODE system, followed by the source identity conditioned motion warping. In Sec.~\ref{sec: train}, we conclude \texttt{DiME}'s entire framework for end-to-end motion transfer.

\vspace{-0.1cm}

\subsection{Motion Transfer via Regularized Optimization}
\label{sec: setup}

\vspace{-0.1cm}
\paragraph{Problem Setup.}
\label{sec: problem}
Let $\mathbb{S} = \mathbb{S}(\mathbf{x}) \subseteq \Omega_{\mathbb{S}}$ denote a source image containing the source object, $\mathbb{R} = \{\mathbb{R}^{(i)}(\mathbf{x}) \subseteq \Omega_{\mathbb{S}}: ~ i = 1,\, \dots, \, N_{\text{ref}}\}$ denote the $N_{\text{ref}}$ reference image(s) containing different views of the same source object.\footnote{In the one-shot (single-view) setup, $N_{\text{ref}} = 0$ and $\mathbb{R} = \emptyset$.} Further, let $\mathbb{D} = \mathbb{D}(\mathbf{x}) \in \{\mathbb{D}(\mathbf{x},\, t) \subseteq~ \Omega_{\mathbb{D}}, \, t = 0,\, 1,\, \dots, \, T\}$ refers to an image from a driving video, which may contain a different object than the source object. Our goal is to estimate $\Phi_{\mathbb{S}\rightarrow\mathbb{D}}$, a dense motion deformation between $\mathbb{S}$ and $\mathbb{D}$, and to generate a series of images $\mathbb{S}\vert_{\mathbb{D}} \in \{\mathbb{S}\vert_{\mathbb{D}}(\mathbf{x},\, t) \subseteq \Omega_{\mathbb{D}}:~ t = 0,\, 1,\, \dots, \, T\}$, where each generated image borrows the \textit{motion} transferred from $\mathbb{D}$ yet \textit{preserves} the original identity of the source object. When $\mathbb{S}$ and $\mathbb{D}$ contains the same identity, the problem equals a video reconstruction task.

Essentially, \texttt{DiME} treats the motion transfer task as a \textit{regularized optimization} problem~\cite{Kantorovitch1958OnTT,benamou2000omt}. While we still aim to minimize the motion differences between the generated $\mathbb{S}\vert_{\mathbb{D}}$ and the driving set $\mathbb{D}$, our approach imposes a continuity constraint that results in a smoother and finer-grained motion deformation field $\Phi_{\mathbb{S} \rightarrow \mathbb{D}}$, the benefit of which is especially evident when there is large motion discrepancy. 
Mathematically, our optimization target writes
\vspace{-0.27cm}
\begin{equation}
    J = d \big(\mathbb{S}\vert_{\mathbb{D}},\, \mathbb{D} \big)\,,
\label{eq: energy}
\end{equation}
\vspace{-0.9cm}
\begin{equation}
s.t.~
    \begin{cases}
      \frac{d \phi_{\mathbb{S} \rightarrow \mathbb{D}}(\epsilon)}{d \epsilon} = f (\phi_{\mathbb{S} \rightarrow \mathbb{D}}(\epsilon),\, \epsilon)\,, \quad \epsilon \in [0,\, 1]\,, \\
      \phi_{\mathbb{S} \rightarrow \mathbb{D}}(0) = \varphi_{\mathbb{S} \rightarrow \mathbb{D}}\,,
    \end{cases}      
\label{eq: constraint}
\vspace{-0.2cm}
\end{equation}

\noindent - Eq.~(\ref{eq: energy}) is the minimization target, $d$ refers to the distance function defined to measure the motion differences between $\mathbb{S}\vert_{\mathbb{D}}$ and $\mathbb{D}$. 
Detailed definitions of $J$ 
is discussed in Sec.~\ref{sec: train}.

\noindent - Eq.~(\ref{eq: constraint}) imposes a continuity constraint on the motion deformation field $\Phi_{\mathbb{S} \rightarrow \mathbb{D}}$, which is essentially an initial value problem which gradually refines a coarse guess ($\varphi_{\mathbb{S} \rightarrow \mathbb{D}}$) of the motion deformation field towards the final prediction ($\phi_{\mathbb{S} \rightarrow \mathbb{D}}(1)$) via $f$, where $\epsilon$ denotes the refinement status. Detailed discussion of the motion evolution is in Sec.~\ref{sec: ode}.

\noindent - The mathematical proof for the existence and uniqueness theorem of the above regularized motion transfer optimization problem is provided in Appendix~\ref{app_sec: ode_supp}.

\begin{figure*}[t]
\centering
\resizebox{0.95\linewidth}{!}{
	\begin{tikzpicture}
		
		\tikzstyle{myarrows}=[line width=0.4mm,draw=blue!50,-triangle 60,postaction={draw, line width=0.05mm, shorten >=0.02mm, -}]
		\tikzstyle{mylines}=[line width=0.4mm,draw=blue!50,]
		\tikzstyle{vecArrow} = [thick, decoration={markings,mark=at position1 with {\arrow[semithick]{open triangle 60}}}, double distance=3pt, shorten >= 5.5pt, preaction = {decorate}, postaction = {draw,line width=3pt, black,shorten >= 4.5pt}]
		
		\pgfmathsetmacro{\dx}{-3.1}
		\pgfmathsetmacro{\dy}{1.4}
		\pgfmathsetmacro{\rectw}{13.41}
		\pgfmathsetmacro{\recth}{10.2}
		 
		\draw[gray,fill=ins_yellow!15, rounded corners, dashed]  (5.8+\dx, -5.45 - \recth/2+\dy) rectangle (5.8+\dx + \rectw, -5.45 + \recth/2+\dy);
		\node at (6.1, 0.7){\textbf{Coarse-to-fine Motion Evolution (Sec.~\ref{sec: ode})}};  
		
		\pgfmathsetmacro{\dx}{0.18}
		\pgfmathsetmacro{\dy}{-0.7}
		\pgfmathsetmacro{\rectw}{10.1}
		\pgfmathsetmacro{\recth}{6.}
		\draw [gray,fill=matcha!20, rounded corners, dashed]  (5.85+\dx, -5.45 - \recth/2+\dy) rectangle (5.85+\dx + \rectw, -5.45 + \recth/2+\dy);
		\node at (14.45, -8.3){\textbf{Source Conditioned}};  
		\node at (14.34, -8.8){\textbf{Warping (Sec.~\ref{sec: ode})}};

		\pgfmathsetmacro{\dx}{18.4}
		\pgfmathsetmacro{\dy}{1}
		
		 
		\pgfmathsetmacro{\x}{-6.9}
		\pgfmathsetmacro{\y}{-0.9}
		\pgfmathsetmacro{\dy}{-0.6}
		\pgfmathsetmacro{\recth}{0.5}
		\pgfmathsetmacro{\rectw}{4.5}
		
		
		
		
		\pgfmathsetmacro{\y}{-0.}
		\node at (-5.1, -0.9+\y){\large$\mathbb{D}$};  
		\node at (-3.6, -0.9+\y) {\includegraphics[width=2.25cm]{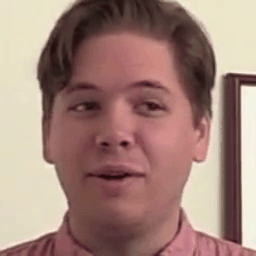}};    
		
		\node at (-5.12, -3.475+\y){\large$\mathbb{S}$};  
		\node at (-3.6, -3.475+\y) {\includegraphics[width=2.25cm]{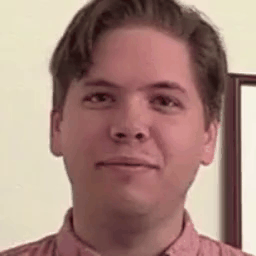}};  
		\draw[gray,fill=black!70] (-3.6,-5.3625+0.35+\y) circle (1.5pt); 
		\draw[gray,fill=black!70] (-3.6,-5.3625+\y) circle (1.5pt); 
		\draw[gray,fill=black!70] (-3.6,-5.3625-0.35+\y) circle (1.5pt);
		\node at (-5.3, -7.25+\y){\large$\mathbb{R}^{(N_{\text{ref}})}$};  
		\node at (-3.6, -7.25+\y) {\includegraphics[width=2.25cm]{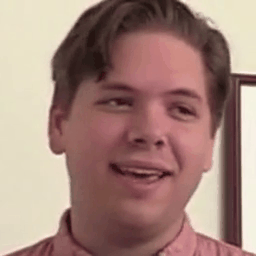}};   
		
		\draw [mylines, color = black!70](-2.3, -3.175+\y) -- (-2, -3.175+\y);  
		\draw [myarrows, color = black!70](-2, -4.75+\y) -- (-1.5, -4.75+\y);  
		\draw [mylines, color = black!70](-2., -4.75+\y) -- (-2, -1.6+\y);  
		\draw [myarrows, color = black!70](-2., -1.6+\y) -- (-1.5, -1.6+\y);  
		\draw [myarrows, color = black!70](-2.3, -1.2+\y) -- (-1.5, -1.2+\y);

		\pgfmathsetmacro{\dx}{0.1}
		\pgfmathsetmacro{\shift}{-0.85}
		\pgfmathsetmacro{\recth}{1.2}
		\pgfmathsetmacro{\rectw}{1.7}
		\pgfmathsetmacro{\textdh}{0.3}

		\pgfmathsetmacro{\start}{2.03}
		\pgfmathsetmacro{\pos}{5}
		\networkLayer{1.6}{0.1}{\start + \shift}{\pos}{color=fb_grey!30}{}
		\networkLayer{1.2}{0.15}{\start + 0.1+ \shift + \dx}{\pos}{color=fb_grey!30}{}
		\networkLayer{0.8}{0.25}{\start + 0.25 + \shift + \dx + \dx}{\pos}{color=fb_grey!30}{}
		\networkLayer{0.4}{0.35}{\start + 0.5 + \shift + \dx + \dx + \dx}{\pos}{color=fb_grey!30}{}
		
		\pgfmathsetmacro{\dx}{0.13}
		\pgfmathsetmacro{\shift}{-2.8}
		\pgfmathsetmacro{\start}{4.4} 
		\networkLayer{0.8}{0.25}{\start + 0.8+ \shift + \dx}{\pos}{color=fb_grey!30}{}
		\networkLayer{1.2}{0.15}{\start + 1.2 + \shift + \dx + \dx}{\pos}{color=fb_grey!30}{}
		\networkLayer{1.6}{0.1}{\start + 1.4 + \shift + \dx + \dx + \dx}{\pos}{color=fb_grey!30}{}
		
		\pgfmathsetmacro{\dx}{0.}
		\pgfmathsetmacro{\dy}{-0.1}
		\draw [decorate,decoration={brace,amplitude=3pt,raise=4ex},line width=1.3pt,color=fb_grey!70](-1.2+\dx,-0.75+\dy) -- (1.8+\dx,-0.75+\dy);
		\node at (0.3+\dx, 0.3+\dy){\textbf{Key-points Extractor}};

		\pgfmathsetmacro{\dx}{1}
		\pgfmathsetmacro{\dy}{-0.1}
		\node at (2.5+\dx, -2.5+\dy){\large $\Delta^{0}$};  
		
		\node at (2.1, -0.8){\large $\{\mathbf{x}^k_{\mathbb{D}}\}$};
		\node at (2.1, -1.5){\large $\{\mathbf{x}^k_{\mathbb{S}}\}$};
		\node at (2.1, -2.2){\large  $\{\mathbf{x}^k_{\mathbb{R}^{(i)}}\}$};
		
		\node at (2.23+\dx, -1.22+\dy){Eq.~(\ref{eq: sparse_motion})};
		
		\draw [myarrows, color = fb_grey](1.7+\dx, -1.6+\dy) -- (2.75+\dx, -1.6+\dy); 
		
		\def\length{sqrt(1+(x-2*y)^2)}
		\begin{axis}[domain=-3:3, view={0}{90},ticks=none,xshift=3.8cm,yshift=-2.75cm, width=2.75cm, height=2.75cm,]
		\addplot3[fb_grey, quiver={u={1/(\length)}, v={(x-y)/(\length)}, scale arrows=0.5}, -stealth,samples=6] {0};
		\draw[gray,fill=fb_grey!70]  (8+\dx, -7.6+\dy - \rectw/2) rectangle (8+\dx + \rectw, -7.6+\dy + \rectw/2);
		\end{axis}
		
		\draw[gray,fill=fb_grey] (4.15+\dx,-1.8+\dy) circle (1.5pt);
		\draw[gray,fill=fb_grey] (4.4+\dx,-1.55+\dy) circle (1.5pt);
		\draw[gray,fill=fb_grey] (4.65+\dx,-1.3+\dy) circle (1.5pt);
		\node[rotate=45] at (4.75+\dx, -1.6+\dy){\footnotesize $(K+1)$};

		\node at (4.1+\dx, -0.75+\dy){\large $\Delta^{K}$};  
		\def\length{sqrt(1+(2*x-y)^4)}
		\begin{axis}[domain=-3:3, view={0}{90},ticks=none,xshift=5.4cm,yshift=-1.cm, width=2.75cm, height=2.75cm,]
		\addplot3[fb_grey, quiver={u={1/(\length)}, v={(x-y)/(\length)}, scale arrows=0.5}, -stealth,samples=6] {0};
		\draw[gray,fill=fb_grey!70]  (8+\dx, -7.6+\dy - \rectw/2) rectangle (8+\dx + \rectw, -7.6+\dy + \rectw/2);
		\end{axis}
		
		
		\pgfmathsetmacro{\dx}{0.7}
		\pgfmathsetmacro{\recth}{1.2}
		\pgfmathsetmacro{\rectw}{2.6}
		\pgfmathsetmacro{\textdh}{0.3} 
		\draw[gray,fill=fb_grey!20, rounded corners, dashed]  (5.75+\dx, -1.6 +\dy - \recth/2) rectangle (5.75+\dx + \rectw, -1.6 +\dy + \recth/2);  
		\node at (5.75+\dx + \rectw/2, -1.6+\dy + \textdh){$\varphi_{\mathbb{S} \rightarrow \mathbb{D}}, \, \varphi_{\mathbb{R}^{(i)} \rightarrow \mathbb{D}}$};
		\node at (5.75+\dx + \rectw/2, -1.6+\dy - \textdh){Eq.~(\ref{eq: initial_dense})}; 
		
		\draw [myarrows, color = fb_grey](5.2+\dx, -1.6+\dy) -- (8.8+\dx, -1.6+\dy); 
		
		
		\pgfmathsetmacro{\recth}{2.25}
		\pgfmathsetmacro{\rectw}{3.8} 
		\draw[gray,fill=ins_yellow!20, rounded corners, dashed]  (9.65, -1.6 - \recth/2) rectangle (9.65 + \rectw, -1.6 + \recth/2);  
		
		\pgfmathsetmacro{\dy}{0.4} 
		 \node at (11.55, -2.35){ODE Solve (Eqs.~(\ref{eq: ode_main}-\ref{eq: ode_initial}))};
  		 \draw[-stealth] (9.65 + 0.1, -2.35+\dy)--(9.65 + \rectw - 0.1,-2.35+\dy); 
		 
   		 \draw plot [smooth] coordinates {(10,-1.6+\dy) (10.25,-1.8+\dy) (10.5, -1.9+\dy)  (10.9, -1.832+\dy) (11.2, -1.6+\dy) (11.5, -1.35+\dy) (11.8, -1.35+\dy) (12., -1.5+\dy) (12.2, -1.58+\dy) (12.5, -1.59+\dy) (13, -1.45+\dy) (13.1, -1.42+\dy)};  
   		 \node (a0) at (10,-1.6+\dy) {};
   		 \draw[fill=red!30] (a0) circle [radius=5pt];       
   		 \node at (10,-1.25+\dy) {\small $\phi(0)$};   
   		 \node (a1) at (10.5,-1.9+\dy) {};
   		 \draw[fill=red!30] (a1) circle [radius=5pt];    
   		 \node (a2) at (11.2,-1.6+\dy) {};
   		 \draw[fill=red!30] (a2) circle [radius=5pt];   
   		 \node (a3) at (11.8,-1.35+\dy) {};
   		 \draw[fill=red!30] (a3) circle [radius=5pt];    
  		  \node (a4) at (12.5,-1.59+\dy) {};
  		  \draw[fill=red!30] (a4) circle [radius=5pt];    
  		  \node (a5) at (13.1,-1.42+\dy) {};
  		  \draw[fill=red!30] (a5) circle [radius=5pt];  
   		 \node at (13.1,-1.78+\dy) {\small $\phi(1)$};   
   		 
		\pgfmathsetmacro{\rectw}{1.7} 
		\draw [myarrows, color = ins_yellow!140](13.5, -1.6) -- (14.25, -1.6); 
		
		\def\length{sqrt(2+(x+1.5*y)^2)}
		\begin{axis}[domain=-3:3, view={0}{90},ticks=none,xshift=14.33cm,yshift=-2.35cm, width=3.25cm, height=3.25cm,]
		\addplot3[ins_yellow, quiver={u={1/(\length)}, v={(x-y)/(\length)}, scale arrows=0.4}, -stealth,samples=10] {0};
		\draw[gray,fill=ins_yellow!20]  (14.25, -1.6 - \rectw/2) rectangle (14.25 + \rectw, -1.6 + \rectw/2);
		\end{axis}
		
		\draw [myarrows, color = ins_yellow!140] (14.25 + \rectw/2, -2.35) -- (14.25 + \rectw/2, -2.94) -- (4.2, -2.94) -- (4.2, -3.85); 
		
		 \node at (14.4, -2.7){Eq.~(\ref{eq: final_dense})};
		\node at (14.1 + \rectw/2, -0.45){$\Phi_{\mathbb{S}\rightarrow\mathbb{D}}, \Phi_{\mathbb{R}^{(i)}\rightarrow\mathbb{D}}$}; 
		
		\pgfmathsetmacro{\dx}{0.}
		\pgfmathsetmacro{\dy}{0.1}
		\draw [decorate,decoration={brace,amplitude=3pt,raise=4ex,mirror}, line width=1.3pt,color = fb_blue!80] (-1.55+\dx,-5.3+\dy) -- (1.3+\dx,-5.3+\dy);
		\node at (-0.13+\dx, -6.35+\dy){\textbf{Feature Extractor}};  
		
		\pgfmathsetmacro{\dx}{0.1}
		\pgfmathsetmacro{\shift}{-0.85}
		
		\pgfmathsetmacro{\start}{5.07}
		\pgfmathsetmacro{\pos}{12.5}
		\networkLayer{2}{0.1}{\start + \shift}{\pos}{color=fb_blue!60}{}
		\networkLayer{1.6}{0.2}{\start + 0.1+ \shift + \dx}{\pos}{color=fb_blue!60}{}
		\networkLayer{1.2}{0.4}{\start + 0.3 + \shift + \dx + \dx}{\pos}{color=fb_blue!60}{}
		\networkLayer{0.8}{0.8}{\start + 0.7 + \shift + \dx + \dx + \dx}{\pos}{color=fb_blue!60}{}

		\pgfmathsetmacro{\recth}{1.2}
		\pgfmathsetmacro{\rectw}{2.7}
		\pgfmathsetmacro{\textdh}{0.3} 
		\node at (1.94, -4.2){\large $\mathbf{F}_{\mathbb{S}}$};
		\node at (2.1, -4.8){\large $\{ \mathbf{F}_{\mathbb{R}^{(i)}}\}$};
		
		\pgfmathsetmacro{\dx}{1.}
		\pgfmathsetmacro{\dy}{0.25}
		\draw[gray,fill=fb_blue!20, rounded corners, dashed]  (1.9+\dx, -4.75 - \recth/2+\dy) rectangle (1.9+\dx + \rectw, -4.75 + \recth/2+\dy);
		\draw [myarrows, color = fb_blue!80](1.7+\dx, -4.75+\dy) -- (2.3+\rectw+\dx, -4.75+\dy); 
		\node at (1.9+\dx + \rectw/2, -4.75 + \textdh+\dy){$\Phi_{\mathbb{S}\rightarrow\mathbb{D}},\, \Phi_{\mathbb{R}^{(i)}\rightarrow\mathbb{D}}$};
		\node at (1.9+\dx + \rectw/2, -4.75+\dy - \textdh){Eq.~(\ref{eq: dense-warped})};
		\pgfmathsetmacro{\dy}{0.}
		\node at (3.14+\dx+\rectw, -4.2+\dy){\large $\mathbf{F}_{\mathbb{S} \rightarrow \mathbb{D}}$};
		\node at (3.3+\dx+\rectw, -4.8+\dy){\large $\{\mathbf{F}_{\mathbb{R}^{(i)} \rightarrow \mathbb{D}} \}$};

		\pgfmathsetmacro{\dx}{2.3}
		\pgfmathsetmacro{\dy}{0.}
		\draw [mylines, color = matcha!150](4.53+\dx, -5.25+\dy) -- (4.53+\dx, -7.55+\dy); 
		\draw [mylines, color = matcha!150](4.53+\dx, -7.55+\dy) -- (4.8+\dx, -7.55+\dy); 
		
		\pgfmathsetmacro{\start}{13.5}
		\pgfmathsetmacro{\pos}{20} 
		\pgfmathsetmacro{\shift}{2.05} 
		\pgfmathsetmacro{\dxl}{0.05} 
		\pgfmathsetmacro{\dxr}{0.13} 
		\networkLayer{1.6}{0.1}{\start + \shift}{\pos}{color=matcha!60}{}
		\networkLayer{1.2}{0.15}{\start + 0. 1+ \shift + \dxl}{\pos}{color=matcha!60}{}
		\networkLayer{0.8}{0.25}{\start + 0.25 + \shift + \dxl*2}{\pos}{color=matcha!60}{}
		\networkLayer{0.4}{0.35}{\start + 0.5 + \shift + \dxl*3}{\pos}{color=matcha!60}{} 
		\networkLayer{0.8}{0.25}{\start + 0.85 + \shift +\dxl*3 + \dxr}{\pos}{color=matcha!60}{}
		\networkLayer{1.2}{0.15}{\start + 1.1 + \shift + \dxl*3 + \dxr*2}{\pos}{color=matcha!60}{}
		\networkLayer{1.6}{0.1}{\start + 1.25 + \shift + \dxl*3 + \dxr*3}{\pos}{color=matcha!60}{}
		
		\draw [myarrows, color = matcha!150](7.1+\dx, -7.55+\dy) -- (7.9+\dx, -7.55+\dy); 
		\draw [decorate,decoration={brace,amplitude=3pt,raise=4ex,mirror}, line width=1.3pt,color = matcha!150] (4.6+\dx,-7.8+\dy) -- (7.3+\dx,-7.8+\dy);
		\node at (6+\dx, -8.8+\dy) {\textbf{Self-flow Predictor}};

		\def\length{sqrt(1+(x-y)^2)}
		\begin{axis}[domain=-3:3, view={0}{90},ticks=none,xshift=10.3cm,yshift=-8.25cm, width=3.25cm, height=3.25cm,]
		\addplot3[matcha, quiver={u={1/(\length)}, v={(x-y)/(\length)}, scale arrows=0.4}, -stealth,samples=10] {0};
		
		\draw[gray,fill=matcha!70]  (8+\dx, -7.6 - \rectw/2+\dy) rectangle (8+\dx + \rectw, -7.6 + \rectw/2+\dy);
		\end{axis}
		
		\draw [myarrows, color = matcha!150](7.45+\dx + \rectw/2, -6.58) -- (7.45+\dx + \rectw/2, -6.35) -- (6.5+\dx + \rectw/2, -6.35) -- (6.5 +\dx+ \rectw/2, -6.35) -- (6.5 +\dx+ \rectw/2, -5.8);


		\pgfmathsetmacro{\dx}{3}
		\pgfmathsetmacro{\dy}{0.25}
		
		\draw [mylines, color = ins_orange!75](5+\dx, -4.75+\dy) -- (5.5+\dx, -4.75+\dy); 
		\draw [mylines, color = ins_orange!75](5.5+\dx, -5.5+\dy) -- (5.5+\dx, -4+\dy); 
		
		\pgfmathsetmacro{\recth}{1.2}
		\pgfmathsetmacro{\rectw}{1.7}

		\pgfmathsetmacro{\rectw}{2.7}
		\draw[gray,fill=matcha!35, rounded corners, dashed]  (5.85+\dx, -5.45 - \recth/2+\dy) rectangle (5.85+\dx + \rectw, -5.45 + \recth/2+\dy);
		\node at (5.85+\dx + \rectw/2, -5.45+\dy + \textdh){$\Phi_{\mathbb{S}\rightarrow\mathbb{S}},\, \Phi_{\mathbb{R}^{(i)}\rightarrow\mathbb{R}^{(i)}}$};
		\node at (5.85+\dx + \rectw/2, -5.45+\dy - \textdh){Eq.~(\ref{eq: appearance-warped})};

		\pgfmathsetmacro{\rectw}{1.7}
		\draw[gray,fill=ins_orange!20, rounded corners, dashed]  (8.9-0.1+\dx, -4.05 - \recth/2+\dy) rectangle (8.9+0.1+\dx + \rectw, -4.05 + \recth/2+\dy);
		\draw[gray,fill=ins_orange!20, rounded corners, dashed]  (8.9-0.1+\dx, -5.45 - \recth/2+\dy) rectangle (8.9+0.1+\dx + \rectw, -5.45 + \recth/2+\dy);
		
		\pgfmathsetmacro{\ddx}{1.1}
		\draw [mylines, color = ins_orange!75](5.5+\dx, -4.05+\dy) -- (9.9+\dx+\ddx, -4.05+\dy); 
		\node at (7.8+\dx+\ddx + \rectw/2, -4.05 + \textdh+\dy){\large$\mathbf{F}_{\text{motion}}$};
		\node at (7.8+\dx+\ddx + \rectw/2, -4.05+\dy - \textdh){Eq.~(\ref{eq: multi_view})};
		
		\node at (7.8+\dx+\ddx + \rectw/2, -5.45+\dy + \textdh){\large$\mathbf{F}_{\text{Id}}$};
		\node at (7.8+\dx+\ddx + \rectw/2, -5.45+\dy - \textdh){Eq.~(\ref{eq: multi_view})};

		\draw [mylines, color = ins_orange!75](5.5+\dx, -5.45+\dy) -- (9.9+\dx+\ddx, -5.45+\dy); 
		\node at (9.9+\dx+\ddx, -4.75+\dy){\Large \color{ins_orange} $\bigoplus$};
		\draw [mylines, color = ins_orange!75](9.9+\dx+\ddx, -4+\dy) -- (9.9+\dx+\ddx, -5.5+\dy); 
		\node at (10.3+\dx+\rectw, -4.45+\dy){\large $\large\mathbf{F}$};
		\node at (10.3+\dx+\rectw, -5.1+\dy){Eq.~(\ref{eq: concat})};
		
		\draw [myarrows, color = ins_orange!75](9.9+\dx+\ddx, -4.75+\dy) -- (12+\dx+\ddx, -4.75+\dy);

		
		\pgfmathsetmacro{\dx}{0.3}
		\pgfmathsetmacro{\shift}{13.4}
		\pgfmathsetmacro{\pos}{12.5}
		
		\networkLayer{0.8}{0.8}{8.1 + \shift}{\pos}{color=ins_orange!50}{}
		\networkLayer{1.2}{0.4}{8.9 + \shift + \dx}{\pos}{color=ins_orange!50}{}
		\networkLayer{1.6}{0.2}{9.3 + \shift + \dx + \dx}{\pos}{color=ins_orange!50}{}
		\networkLayer{2}{0.1}{9.5 + \shift + \dx + \dx + \dx}{\pos}{color=ins_orange!50}{}  
		
		\pgfmathsetmacro{\dx}{5.6}
		
		\draw [decorate,decoration={brace,amplitude=3pt,raise=4ex,mirror}, line width=1.3pt,color = ins_orange!75] (10.6+\dx,-5.3+\dy) -- (13.4+\dx,-5.3+\dy);
		\node at (12+\dx, -6.35+\dy){\textbf{Image Generator}};   
		
		\draw [myarrows, color = ins_orange!75](13.1+\dx, -4.75+\dy) -- (13.85+\dx, -4.75+\dy); 
		 
		\node at (15.1+\dx, -4.75+\dy) {\includegraphics[width=2.25cm]{fig/dfdc/ahslbdxbjz/178}};  
		\node at (15.1+\dx, -3.25+\dy){\large$\mathbb{S}\vert_{\mathbb{D}}$}; 
		
		\node at (13.55+\dx, -1.+\dy){\large \textbf{Optimization Target (Eq.~(\ref{eq: energy})):}}; 
		\node at (13.35+\dx, -1.75+\dy){\large $J = d \big(\mathbb{S}\vert_{\mathbb{D}},\, \mathbb{D} \big)$ s.t. Eq.~(\ref{eq: constraint})};

    	 \end{tikzpicture}} 
	\vspace{-0.2cm}
\caption{Overview of \texttt{DiME}'s framework~(Sec.~\ref{sec: overview}). An encoder first extracts image features of the input source and reference images. A dense motion deformation field from the source to the driving domain is predicted and applied to warp the source features to the driving domain, conditioned on the source identity~(Sec.~\ref{sec: ode}). Finally, an image generator takes the driving-domain-warped features and outputs the final prediction results of transferring the motion from the driving videos to the source identity. In summary, \texttt{DiME} is trained in an end-to-end fashion and optimized with the introduced continuity constraints on the coarse-to-fine evolution of the motion deformation~(Sec.~\ref{sec: train}).}
\label{fig: fw}
\end{figure*}

\vspace{-0.4cm}
\paragraph{Workflow Overview.}
\label{sec: overview}
In general, three sequential steps form our end-to-end \texttt{DiME} framework (Fig.~\ref{fig: fw}): \hypertarget{step-1}{1)} An encoder-structured network first extracts features of the source identity ($\mathbf{F}_{\mathbb{S}}$) from the input source ($\mathbb{S}$) and reference images ($\mathbb{R}$) respectively; \hypertarget{step-2}{2)} A dense motion deformation field from the source image to the driving domain is predicted, and then applied to warp the extracted source features ($\mathbf{F}_{\mathbb{S}}$) to the driving domain conditioned on the source identity; \hypertarget{step-3}{3)} A following decoder-structured network takes the driving-domain-warped features ($\mathbf{F}$) and generates $\mathbb{S}\vert_{\mathbb{D}}$, the final prediction results of transferring the motion from the driving videos to the source identity.


\subsection{Differential Coarse-to-fine Motion Evolution}
\label{sec: ode}
\vspace{-0.1cm}
In this section, we present the key component of \texttt{DiME}, the coarse-to-fine motion deformation field estimation. Briefly, we solve the motion deformation prediction through an initial value problem (IVP), which initializes an ODE with a coarse estimation, followed by a gradual refinement towards a final fine-grained prediction~(Eq.~(\ref{eq: constraint})).


\vspace{-0.35cm}

\paragraph{Coarse motion initialization.}
\label{sec: keypoint}

To start with, \texttt{DiME} uses an encoder-decoder network~\cite{Siarohin_2019_CVPR,Siarohin_2019_NeurIPS,Wang_2021_CVPR} to extract $K$ corresponding key-points from the source ($\mathbb{S}$) and reference images ($\mathbb{R}$) of the source object, as well as from the driving object ($\mathbb{D}$), respectively,
\vspace{-0.13cm}
\begin{equation}
\{\mathbf{x}^{k}_{\mathbb{S}}\},\,
\{\mathbf{x}^{k}_{\mathbb{R}^{(i)}}\},\, \{\mathbf{x}^{k}_{\mathbb{D}}\},\quad (k = 1,\, \dots,\, K,\,\, i = 1,\, \dots,\, N_{\text{ref}}).
\label{eq: keypoints}
\vspace{-0.17cm}
\end{equation}
The keypoint-wise displacements from the source domains ($\mathbb{S}$, $\mathbb{R}$) to the driving domain ($\mathbb{D}$) are therefore:
\vspace{-0.15cm}
\begin{equation}
\{\Delta^{0},\, \Delta^{k}_{\mathbb{S}\rightarrow\mathbb{D}} = \mathbf{x}^{k}_{\mathbb{S}} - \mathbf{x}^{k}_{\mathbb{D}}\}, \,\,
\{\Delta^{0},\, \Delta^{k}_{\mathbb{R}^{(i)}\rightarrow\mathbb{D}} = \mathbf{x}^{k}_{\mathbb{S}} - \mathbf{x}^{k}_{\mathbb{D}}\},
\label{eq: sparse_motion}
\vspace{-0.2cm}
\end{equation}
$\Delta^{0}:=\mathbf{0}$ is an additional term for the static background. A convolutional network is then applied to regress over the sparse, keypoint-wise displacements, to obtain the initial coarse predictions on the dense motion transformation:
\vspace{-0.25cm}
\begin{equation}
\varphi_{\mathbb{S} \rightarrow \mathbb{D}}, \,\, \varphi_{\mathbb{R}^{(i)} \rightarrow \mathbb{D}},\quad (i = 1,\, \dots,\, N_{\text{ref}}).
\label{eq: initial_dense}
\vspace{-0.15cm}
\end{equation}


\vspace{-0.35cm}

\paragraph{Motion deformation refinement in an ODE system.}
\label{sec: refine}

Given the initial coarse predictions on the motion deformation in Eq.~(\ref{eq: initial_dense}), \texttt{DiME} gradually refines the coarse motion deformation fields in a system of non-linear ODEs. Specifically,  ($\epsilon \in [0,\, 1]$, $i = 1,\, \dots,\, N_{\text{ref}}$)
\vspace{-0.2cm}
\begin{equation}
\begin{cases}
    \frac{d \phi_{\mathbb{S} \rightarrow \mathbb{D}}(\epsilon)}{d \epsilon} = f_0 (\phi_{\mathbb{S} \rightarrow \mathbb{D}}(\epsilon),\, \epsilon)\,, \\ \vspace{-0.1cm}
    \frac{d \phi_{\mathbb{R}^{(i)} \rightarrow \mathbb{D}}(\epsilon)}{d \epsilon} = f_i (\phi_{\mathbb{R}^{(i)} \rightarrow \mathbb{D}}(\epsilon),\, \epsilon)\,,
\label{eq: ode_main}
\end{cases}
\end{equation}
\vspace{-0.1cm}
\begin{equation}
s.t.~
    \begin{cases}
      \phi_{\mathbb{S} \rightarrow \mathbb{D}}(0) = \varphi_{\mathbb{S} \rightarrow \mathbb{D}}\,, \\
      \phi_{\mathbb{R}^{(i)} \rightarrow \mathbb{D}}(0)) = \varphi_{\mathbb{R}^{(i)} \rightarrow \mathbb{D}}\,,
\label{eq: ode_initial}
\end{cases}
\end{equation}
where $f_0,\, f_1,\, \dots,\, f_{N_{\text{ref}}}$ are convolutional networks representing the motion evolution dynamics.\footnote{Note when we perform one-shot motion transfer, no reference image is needed, and thus the above ODE system, Eqs.~(\ref{eq: ode_main}-\ref{eq: ode_initial}), reduces to a single ODE with an initial condition.} (See the proof for the existence and uniqueness theorem of solutions to the above non-linear ODE system in Appendix~\ref{app_sec: ode_supp}.)

Essentially, the integration of the above ODE system (Eqs.~(\ref{eq: ode_main}-\ref{eq: ode_initial})) equals a refinement process from the initial coarse motion deformations (Eq.~(\ref{eq: initial_dense})) to the final result of the fine-grained motion deformation estimation:
\vspace{-0.3cm}
\begin{equation}
\begin{cases}
\Phi_{\mathbb{S} \rightarrow \mathbb{D}} = \phi_{\mathbb{S} \rightarrow \mathbb{D}}(1)\,, \\ 
\Phi_{\mathbb{R}^{(i)} \rightarrow \mathbb{D}} = \phi_{\mathbb{R}^{(i)} \rightarrow \mathbb{D}}(1)\,, \quad i = 1,\, \dots,\, N_{\text{ref}}\,.
\label{eq: final_dense}
\end{cases}
\vspace{-0.25cm}
\end{equation}

In Monkey-Net~\cite{Siarohin_2019_CVPR}, the initial regression result $\varphi_{\mathbb{S}\rightarrow\mathbb{D}}$ is directly used to warp the source image $\mathbb{S}$ and generate the final image prediction. In FOMM~\cite{Siarohin_2019_NeurIPS}, additional first-order displacements are applied for predicting the regression coefficients. In~\cite{Siarohin_2021_CVPR}, the sparse displacements are alternatively formulated via principal component analysis (PCA). \texttt{DiME}'s motion evolution approach is a generalized model applicable for all the above approaches. Without the motion evolution, \texttt{DiME} reduces to~\cite{Siarohin_2019_CVPR,Siarohin_2019_NeurIPS,Siarohin_2021_CVPR}, where the initial status $\varphi_{\mathbb{S}\rightarrow\mathbb{D}}$ formulated in the corresponding work is directly used as the final prediction for $\Phi_{\mathbb{S}\rightarrow\mathbb{D}}$.


\vspace{-0.4cm}

\paragraph{Source identity conditioned motion warping.}
\label{sec: appearance}

Upon obtaining the refined dense motion deformation $\Phi_{\mathbb{S}\rightarrow\mathbb{D}}$, we have the driving-domain-warped features:
\vspace{-0.3cm}
\begin{equation} 
\begin{cases}
\mathbf{F}_{\mathbb{S}\rightarrow\mathbb{D}} = \Phi_{\mathbb{S}\rightarrow\mathbb{D}} \circ \mathbf{F}_{\mathbb{S}}\,, \\ 
\mathbf{F}_{\mathbb{R}^{(i)}\rightarrow\mathbb{D}} = \Phi_{\mathbb{R}^{(i)}\rightarrow\mathbb{D}} \circ \mathbf{F}_{\mathbb{R}^{(i)}} \,, \quad i = 1,\, \dots,\, N_{\text{ref}}\,.
\label{eq: dense-warped}
\end{cases}
\vspace{-0.2cm}
\end{equation}
However, the pixel-wise domain deformation warping is not enough~(Fig.~\ref{fig: dfdc_recon}). In reality, large deformation resulting from significant pose changes are common, e.g., head turning right from left, eyes opening and closing, etc. Without corresponding poses in the source and references, 
it may not be recoverable by the domain deformation in Eq.~(\ref{eq: final_dense}) irregardless of its refinement level, i.e., one cannot deform intensities from a non-existing domain to another. That been said, for domains with \textit{symmetrical} structures, especially in the case of human faces and bodies, features in the existing regions of the source domain itself could be potentially \textit{re-used} to assist $\Phi_{\mathbb{S}\rightarrow\mathbb{D}}$ in warping the motion from the source to the driving domain, and inpainting the missing regions, by conditioning on flow fields extracted from the source identity.

Specifically, \texttt{DiME} applies an encoder-decoder network~\cite{Zhou_2016_ECCV} to predict the ``self-flow'' --- flow deformation fields ($\Phi_{\mathbb{S}\rightarrow\mathbb{S}},\, \Phi_{\mathbb{R}^{(i)}\rightarrow\mathbb{R}^{(i)}}$) that borrows features from the existing regions of the source domain \textit{itself} and warps them to the missing regions on the driving domain:
\vspace{-0.25cm}
\begin{equation} 
\hspace*{-0.15cm}
\begin{cases}
\mathbf{F}_{\mathbb{S}\rightarrow\mathbb{S}} = \Phi_{\mathbb{S}\rightarrow\mathbb{S}} \circ \mathbf{F}_{\mathbb{S} \rightarrow \mathbb{D}}\,, \\ 
\mathbf{F}_{\mathbb{R}^{(i)}\rightarrow\mathbb{R}^{(i)}} = \Phi_{\mathbb{R}^{(i)}\rightarrow\mathbb{R}^{(i)}} \circ \mathbf{F}_{\mathbb{R}^{(i)} \rightarrow \mathbb{D}} \,, \,\, i = 1,\, \dots,\, N_{\text{ref}}\,. \hspace*{-0.1cm}
\label{eq: appearance-warped}
\end{cases}
\vspace{-0.1cm}
\end{equation}
In this way, $\mathbf{F}_{\mathbb{S}\rightarrow\mathbb{S}}$ ($\mathbf{F}_{\mathbb{R}^{(i)}\rightarrow\mathbb{R}^{(i)}}$) learns where to copy features from the source domain to the missing regions on the driving domain (if similar parts exist), and therefore helps to reduce the burden of generating all the missing regions from scratch in the image generator.

Finally, we apply a soft confidence mask $\alpha^{(i)} (i=0,\, ...,\, N_{\text{ref}})$ to perform a weighted-sum over the warped features from the source ($\mathbb{S}$) and reference images ($\mathbb{R}$), 
\vspace{-0.3cm}
\begin{equation}
\begin{cases}
\mathbf{F}_{\text{motion}} = \alpha^{(0)} \mathbf{F}_{\mathbb{S} \rightarrow \mathbb{D}} + \sum\limits_{i=1}^{N_{\text{ref}}} \, \alpha^{(i)} \mathbf{F}_{\mathbb{R}^{(i)}\rightarrow\mathbb{D}}\,, \\ 
\mathbf{F}_{\text{Id}} = \alpha^{(0)} \mathbf{F}_{\mathbb{S} \rightarrow \mathbb{S}} + \sum\limits_{i=1}^{N_{\text{ref}}} \, \alpha^{(i)} \mathbf{F}_{\mathbb{R}^{(i)}\rightarrow\mathbb{R}^{(i)}}\,,
\end{cases}
\label{eq: multi_view}
\vspace{-0.2cm}
\end{equation}
and condition the source identity warped feature ($\mathbf{F}_{\text{Id}}$) on the motion deformation warped features~($\mathbf{F}_{\text{motion}}$), to obtain $\mathbf{F}$, for the final motion-transferred image generation~(Fig.~\ref{fig: fw}):
\vspace{-0.2cm}
\begin{equation}
\mathbf{F} = \mathbf{F}_{\text{motion}} \oplus \mathbf{F}_{\text{Id}}\,.
\label{eq: concat}
\vspace{-0.2cm}
\end{equation}
$\mathbf{F}$ is then forwarded to an image generator~\cite{Siarohin_2019_CVPR,Siarohin_2019_NeurIPS,Wang_2021_CVPR} to obtain the final output of the generated image $\mathbb{S}\vert_{\mathbb{D}}$.


\subsection{End-to-End Motion Transfer Optimization}
\label{sec: train}

\vspace{-0.1cm}
As mentioned in Sec.~\ref{sec: problem}, during the end-to-end training of \texttt{DiME}, we aim to minimize the motion differences between $\mathbb{S}\vert_{\mathbb{D}}$ and $\mathbb{D}$, while encouraging $\mathbb{S}\vert_{\mathbb{D}}$ to preserve the source identity. Following previous work~\cite{Siarohin_2019_NeurIPS,Siarohin_2021_CVPR}, we define $J$ in Eq.~(\ref{eq: energy}) as the multi-resolution perceptual reconstruction metric between the generated $\mathbb{S}\vert_{\mathbb{D}}$ and the target $\mathbb{D}$, which is computed from their corresponding feature spaces from the pre-trained \texttt{VGG-19}~\cite{Johnson_2016_ECCV} network:
\vspace{-0.25cm}
\begin{equation}
J = \sum_{i=1}^5 \sum_{j=1}^4 \big\vert V_i (\mathbb{S}_{\mathbb{D},\, j}) - V_i (\mathbb{D}_j) \big\vert + \lambda R, \quad \lambda > 0\,,
\vspace{-0.25cm}
\end{equation}
where $i$ denotes the $i^{\text{th}}$ layer of the \texttt{VGG-19}, and $j$ refers to the $j^{\text{th}}$ downsampling level of corresponding images, and $R$ is an additional regularization term which encourages the key-points extraction consistency: 
\vspace{-0.25cm}
\begin{equation}
R = \sum\limits_{k=1}^{K} \big\vert \chi \circ \mathbf{x}^k_{\mathbb{S}} - \mathbf{x}^k_{\chi \circ \mathbb{S}} \big\vert,
\vspace{-0.25cm}
\end{equation}
$\chi$ is certain random geometric transformations~\cite{Jakab_2018_Neurips,Siarohin_2019_NeurIPS}, minimizing $R$ encourages the equivalence of extracted key-points ($\mathbf{x}^k_{\mathbb{S}},\, \mathbf{x}^k_{\chi \circ \mathbb{S}}$) under different geometric transformations.

\vspace{-0.1cm}

\section{Experiments}
\label{sec: exp}






\begin{table*}[h] 
\centering
\renewcommand{\arraystretch}{1.25}
\resizebox{0.95\linewidth}{!}{
 
\begin{tabular}{cccccccccc}
\specialrule{.15em}{.05em}{.05em}

\multicolumn{1}{c}{\multirow{2}{*}{\textbf{\textit{Dataset}}}} & \multicolumn{1}{c}{\multirow{2}{*}{\textbf{\textit{Method}}}} & \multicolumn{6}{c}{\textbf{\textit{Video Reconstruction}}} & \multicolumn{2}{c}{\textbf{\textit{Image Animation}}} \\

\cmidrule(lr){3-8}
\cmidrule(lr){9-10}

&  & \multicolumn{1}{c}{\multirow{1}{*}{\texttt{L1} ($\downarrow$)}} & \multicolumn{1}{c}{\multirow{1}{*}{\texttt{LPIPS} ($\downarrow$)}} & \multicolumn{1}{c}{\multirow{1}{*}{\texttt{FID} ($\downarrow$)}} & \multicolumn{1}{c}{\multirow{1}{*}{\texttt{MS-/SSIM} ($\uparrow$)}} & \multicolumn{1}{c}{\multirow{1}{*}{\texttt{PSNR} ($\uparrow$)}} & \multicolumn{1}{c}{\multirow{1}{*}{\texttt{AKD} ($\downarrow$)}} & {\multirow{1}{*}{\texttt{FID} ($\downarrow$)}} & \multicolumn{1}{c}{\multirow{1}{*}{\texttt{CSIM} ($\uparrow$)}} \\

\specialrule{.15em}{.05em}{.05em}

\multicolumn{1}{c}{\multirow{10}{*}{\shortstack{\texttt{VoxCeleb}\\(507)}}} &  \texttt{FOMM} \cite{Siarohin_2019_NeurIPS} & 0.047 (0.040) & 0.14 (0.11)  & 50.96 (43.28) & 0.84/0.75 (0.87/0.78) & 22.74 (23.77) & 2.29 (1.36) & 62.88 (52.10) & 0.78 (0.81) \\ 

& \texttt{AA} \cite{Siarohin_2021_CVPR} & 0.044 (0.034)  & 0.13 (0.18) & 50.69 (44.02) & 0.86/0.77 (0.90/0.82) & 23.18 (24.61) & 2.15 (1.26) & 86.03 (79.59) & 0.75 (0.75) \\ 

& \texttt{X2Face} \cite{Wiles2018X2FaceAN} & 0.078 & N/A  & N/A & N/A & N/A & 7.69 & N/A & N/A  \\ 

& \texttt{MonkeyNet} \cite{Siarohin2018AnimatingAO} & 0.049 & 0.136  & N/A & N/A & N/A & 1.39 & N/A & N/A  \\ 

& \texttt{f-v2v} \cite{Wang_2021_CVPR} & 0.042 & N/A  & 69.13 & 0.85/0.80 & 24.37 & 2.07 & 55.64 & 0.75  \\ 

& \texttt{LIA} \cite{Wang2022LatentIA} & 0.041 & 0.123  & N/A & N/A & N/A & 1.31 & N/A & N/A  \\ 

& \texttt{SAMT} \cite{Tao2022StructureAwareMT} & 0.030 & N/A  & N/A & N/A & N/A & 1.24 & N/A & N/A  \\ 

& \texttt{TPSMM} \cite{Zhao2022ThinPlateSM} & 0.045 &   N/A  & N/A & N/A & N/A & 1.22 & N/A & N/A  \\ 

& \texttt{IAPM} \cite{Shalev2020ImageAW} & 0.034 &   N/A  & N/A & N/A & N/A & 1.33 & N/A & N/A  \\ 

& \texttt{DiME} & {\bf{0.027}}   & {\bf{0.070}}  & {\bf{25.43}} & {\bf{0.93}}/{\bf{0.86}} & {\bf{26.20}} & {\bf{1.18}} & {\bf{47.21}} & {\bf{0.90}}  \\ 
  
\hline

\multicolumn{1}{c}{\multirow{3}{*}{\shortstack{\texttt{DFDC}\\(1807)}}} &  \texttt{FOMM} \cite{Siarohin_2019_NeurIPS} & 0.034 (0.029) & 0.140 (0.119) & 61.59 (51.59) & 0.86/0.80 (0.89/0.82) & 25.65 (26.72)  &  2.81 (2.42) & 84.18 (78.76) & 0.81 (0.85) \\ 

& \texttt{AA} \cite{Siarohin_2021_CVPR} & 0.039 (0.035) & 0.156 (0.140) & 89.88 (83.79)  & 0.86/0.79 (0.89/0.80)  & 23.64 (24.21) & 2.52 (2.43) & 172.35 (156.32) & 0.73 (0.74) \\  

& \texttt{DiME} & {\bf{0.018}}  & {\bf{0.0061}} & {\bf{33.22}} & {\bf{0.95}}/{\bf{0.91}} & {\bf{27.99}} & {\bf{1.28}}  & {\bf{64.89}} & {\bf{0.92}}   \\

\hline

\multicolumn{1}{c}{\multirow{3}{*}{\shortstack{\texttt{UvA-Nemo}\\(124)}}} &  \texttt{FOMM} \cite{Siarohin_2019_NeurIPS} & 0.0139 (0.0120) & 0.034 (0.027) & 19.60 (18.03) & 0.96/0.95 (0.97/0.95) & 32.90 (33.91) & 1.17 (1.12) & 27.46 (23.80) & 0.93 (0.94) \\ 

& \texttt{AA} \cite{Siarohin_2021_CVPR} & 0.0169 (0.0151) & 0.041 (0.037) & 30.09 (40.31) & 0.95/0.93 (0.96/0.94) & 30.55 (31.00) & 1.32 (1.24) & 72.11 (67.65) & 0.83 (0.83) \\  

& \texttt{DiME} & {\bf{0.0061}} & {\bf{0.013}} & {\bf{10.35}} & {\bf{0.99}}/{\bf{0.98}} & {\bf{39.40}} & {\bf{0.87}} & {\bf{23.46}} & {\bf{0.98}} \\

\specialrule{.08em}{.05em}{.05em}
\specialrule{.08em}{.05em}{.05em}

\multicolumn{1}{c}{\multirow{5}{*}{\shortstack{\texttt{TED-talks}\\(120)}}} &  \texttt{FOMM} \cite{Siarohin_2019_NeurIPS} & 0.33 (0.030) & 0.18 (0.16) & 56.57 (50.93) & 0.78/0.74 (0.83/0.81) & 25.11 (25.95) & 6.99 (6.87) & 123.31 (112.75) & 0.92 (0.93) \\ 

& \texttt{AA} \cite{Siarohin_2021_CVPR} & 0.025 (0.023) & 0.15 (0.14) & 46.86 (42.97) & 0.82/0.79 (0.85/0.82) & 25.27 (26.05) & 3.71 (3.12) & 112.35 (109.70) & 0.93 (0.95) \\ 

& \texttt{LIA} \cite{Wang2022LatentIA} & 0.027 & 0.11 & N/A & N/A & N/A & 3.14 & N/A & N/A \\  

& \texttt{TPSMM} \cite{Zhao2022ThinPlateSM} & 0.027 & N/A & N/A & N/A & N/A & 3.39 & N/A & N/A \\ 

& \texttt{DiME} & {\bf{0.018}} & {\bf{0.11}} & {\bf{31.84}} & {\bf{0.89}}/{\bf{0.86}} & {\bf{28.15}} & {\bf{2.97}} & {\bf{93.55}} & {\bf{0.95}} \\ 
 
\hline

\multicolumn{1}{c}{\multirow{3}{*}{\shortstack{\texttt{AIST}\\(493)}}} &  \texttt{FOMM} \cite{Siarohin_2019_NeurIPS} & 0.015 (0.013)  & 0.056 (0.048) & 63.17 (49.33) & 0.94/0.94 (0.95/0.94) & 26.27 (26.92) & 6.69 (5.47)  & 114.57 (109.81) & 0.99 (0.99)  \\ 

& \texttt{AA} \cite{Siarohin_2021_CVPR} & 0.029 (0.028) & 0.239 (0.234) & 99.95 (91.93) & 0.94/0.91 (0.95/0.92) & 20.40 (20.54) & 5.67 (5.09) & 198.43 (182.63) & 0.98 (0.99) \\  

& \texttt{DiME} & {\bf{0.011}} & {\bf{0.043}} & {\bf{37.71}} & {\bf{0.96}}/{\bf{0.95}} & {\bf{26.97}} & {\bf{3.66}}  & {\bf{105.20}} & {\bf{0.99}}    \\ 
  
\hline

\multicolumn{1}{c}{\multirow{9}{*}{\shortstack{\texttt{Tai-Chi-HD}\\(87)}}} &  \texttt{FOMM} \cite{Siarohin_2019_NeurIPS}  & 0.057 (0.046) & 0.24 (0.18) & 123.78 (95.95) & 0.69/0.64 (0.77/0.72) & 20.45 (21.73) & 9.69 (7.87) & 148.86 (134.46) & 0.99 (0.99) \\ 

& \texttt{AA} \cite{Siarohin_2021_CVPR} & 0.059 (0.053) & 0.21 (0.17) & 109.56 (86.57) & 0.70/0.57 (0.76/0.63) & 20.31 (21.19) & 7.30 (6.67) & 159.54 (147.73) & 0.99 (0.99) \\  

& \texttt{X2Face} \cite{Wiles2018X2FaceAN} & 0.080 & N/A  & N/A & N/A & N/A & 17.65 & N/A & N/A  \\ 

& \texttt{MonkeyNet} \cite{Siarohin2018AnimatingAO} & 0.077 & N/A  & N/A & N/A & N/A & 10.79 & N/A & N/A  \\ 

& \texttt{LIA} \cite{Wang2022LatentIA} & 0.057 & 0.18 & N/A & N/A & N/A & 4.57 & N/A & N/A \\

& \texttt{SAMT} \cite{Tao2022StructureAwareMT} & 0.030 & N/A  & N/A & N/A & N/A & 1.24 & N/A & N/A  \\ 

& \texttt{TPSMM} \cite{Zhao2022ThinPlateSM} & 0.047 & N/A & N/A & N/A & N/A & {\bf{4.25}} & N/A & N/A \\

& \texttt{IAPM} \cite{Shalev2020ImageAW} & 0.034 &   N/A  & N/A & N/A & N/A & 1.33 & N/A & N/A  \\ 

& \texttt{DiME}  & {\bf{0.035}} & {\bf{0.15}} & {\bf{76.41}} & {\bf{0.83}}/{\bf{0.78}} & {\bf{23.08}} & 5.36 & {\bf{123.10}} & {\bf{0.99}} \\ 
  
\hline

\multicolumn{1}{c}{\multirow{3}{*}{\shortstack{\texttt{Fashion}\\(100)}}} &  \texttt{FOMM} \cite{Siarohin_2019_NeurIPS} & 0.013 (0.013) & 0.033 (0.033) & 30.73 (30.18) & 0.96/0.93 (0.95/0.93) & 27.06 (26.12) & 1.09 (1.08) & 41.10 (39.52) & 0.99 (0.99) \\ 

& \texttt{AA} \cite{Siarohin_2021_CVPR} & 0.015 (0.014) & 0.040 (0.039) & 37.99 (36.10) & 0.95/0.92 (0.96/0.92) & 25.88 (25.98) & 1.54 (1.50) & 78.01 (70.76) & 0.99 (0.99) \\  

& \texttt{DiME} & {\bf{0.010}} & {\bf{0.030}} & {\bf{22.25}} & {\bf{0.97}}/{\bf{0.95}} & {\bf{26.95}} & {\bf{1.00}} & {\bf{32.64}} & {\bf{0.99}} \\

\specialrule{.08em}{.05em}{.05em}
\specialrule{.08em}{.05em}{.05em}

\multicolumn{1}{c}{\multirow{3}{*}{\shortstack{\texttt{Bair}\\(128)}}} & \texttt{FOMM} \cite{Siarohin_2019_NeurIPS} & 0.030 (0.024) & 0.067 (0.047) & 107.70 (45.41) & 0.91/0.88 (0.94/0.91) & 23.64 (25.35) & N/A & 107.70 (114.54) & N/A \\ 

& \texttt{AA} \cite{Siarohin_2021_CVPR} & 0.052 (0.048) & 0.137 (0.121) & 217.87 (77.14) & 0.87/0.76 (0.90/0.77) & 19.75 (19.75)  & N/A & 217.87 (201.78) & N/A\\ 

& \texttt{DiME} & {\bf{0.018}} & {\bf{0.027}} & {\bf{34.44}} & {\bf{(0.97) 0.94}} & {\bf{27.75}} & N/A & {\bf{96.91}} & N/A \\

\specialrule{.08em}{.05em}{.05em}
\specialrule{.08em}{.05em}{.05em}
 
\multicolumn{1}{c}{\multirow{3}{*}{\shortstack{\texttt{MGif}\\(100)}}} & \texttt{FOMM} \cite{Siarohin_2019_NeurIPS} & 0.033 (0.031) & 0.112 (0.106) & 154.88 (146.46) & 0.87/0.84 (0.87/0.84) & 19.56 (19.79) & N/A & 196.94 (183.68) & N/A \\ 

& \texttt{AA} \cite{Siarohin_2021_CVPR} & 0.026 (0.026) & 0.092 (0.089) & 136.84 (131.89) & 0.90/0.85 (0.90/0.86) & 20.22 (20.35) & N/A & 331.72 (315.71) & N/A \\ 

& \texttt{DiME} & {\bf{0.021}} & {\bf{0.072}} & {\bf{117.74}} & {\bf{(0.92) 0.90}} & {\bf{21.34}} & N/A & {\bf{180.30}} & N/A \\ 

\specialrule{.15em}{.05em}{.05em}
 
\end{tabular} 
}

\vspace{-0.2cm}

\caption{Comparing \texttt{DiME} with the state-of-the-art methods on human face datasets (\texttt{VoxCeleb}, \texttt{DFDC}, \texttt{UvA-Nemo}), human body datasets (\texttt{TED-talks}, \texttt{AIST}, \texttt{Tai-Chi-HD, \texttt{Fashion}}), robot dataset (\texttt{Bair}) and cartoon animal dataset (\texttt{MGif}). The number of reconstruction testing samples are listed in parentheses below each dataset name. For animation tests, 200 source-driving object pairs are randomly generated for each dataset. For fairer comparisons, we additionally list, in the parentheses, the \textit{best} scores of \texttt{FOMM} \cite{Siarohin_2019_NeurIPS} and Articulated Animation (\texttt{AA}) \cite{Siarohin_2021_CVPR} computed from the \textit{same} reference images used by \texttt{DiME}. \texttt{DiME} consistently achieves the best performance on \textit{all} metrics across \textit{all} nine datasets.
}
\label{tab: all}
\end{table*}

\vspace{-0.1cm}

We evaluate the performance of \texttt{DiME} over three different tasks: (1) Video reconstruction~(Sec.~\ref{sec: reconstruction}), where the source image and driving video contains the same object identity; (2) Image animation~(Sec.~\ref{sec: animation}), where the source image and driving video hold different identities; (3) Model generality~(Sec.~\ref{sec: generality}), where we test the models on videos containing \textit{novel} object categories that are not seen during training. 
We compare \texttt{DiME} with three current state-of-the-art methods: \texttt{FOMM} \cite{Siarohin_2019_NeurIPS}, Articulated Animation (\texttt{AA}) \cite{Siarohin_2021_CVPR} and \texttt{face-vid2vid} \cite{Wang_2021_CVPR}. We demonstrate that \texttt{DiME} significantly outperforms the state-of-the-arts across different datasets with varying object categories~(Sec.~\ref{sec: comparisons}), and generalizes well to a wide range of out-of-domain datasets~(Sec.~\ref{sec: generality}).

\vspace{-0.4cm}

\paragraph{Datasets}
\label{sec: datasets}

\vspace{-0.1cm}

We experiment on nine datasets listed as below, with object categories varying from human faces, human bodies, talking head, to robots and cartoons.

\noindent - \texttt{VoxCeleb} \cite{Nagrani_2017}: a talking head video dataset from \texttt{YouTube}. We used the same pre-processing and train-test splitting strategies as \cite{Siarohin_2019_NeurIPS}. We obtained 18,314 videos for training and 507 for testing, with lengths varying from 64 to 1,024 frames. All videos were resized to $256 \times 256$. 

\noindent -  \texttt{DFDC} \cite{DFDC_2020}: the Deepfake Detection Challenge dataset with 15,039 videos for training and 1,807 for testing, each with 300 frames and were processed to contain only the talking-heads with the same pipeline for \texttt{VoxCeleb}. 

\noindent - \texttt{UvA-Nemo} \cite{Hamdi_2015_Tran_MM,Hamdi_2012_ECCV}: a facial analysis dataset containing 1,110 videos for training and 124 for testing, each starts from a neutral expression~\cite{Siarohin_2019_NeurIPS,Wang_2018_CVPR}.

\noindent - \texttt{TED-talks} \cite{Siarohin_2021_CVPR}: a \texttt{TED} talk video dataset of 1,200 videos for training and 120 for testing, with lengths from 64 to 1,024 frames. We cropped the upper part of the human body, and downscaled all videos to $384 \times 384$~\cite{Siarohin_2021_CVPR}.

\noindent - \texttt{AIST Dance DB} \cite{AIST_2019}: a dance dataset containing 12,187 videos for training and 493 for testing, with lengths from 400 to 1,800 frames. Videos were cropped according to body tracking\footnote{Code in \url{https://github.com/ITCoders/Human-detection-and-Tracking}} and resized to $256 \times 256$. 

\noindent - \texttt{Tai-Chi-HD} \cite{Nagrani_2017}: a dataset with 769 tai-chi videos from \texttt{YouTube} following \cite{Nagrani_2017}. Overall, 682 videos were used for training and 87 for testing.

\noindent - \texttt{Fashion} \cite{Zablotskaia_2019}: a dataset of 600 videos (500 for training, 100 for testing), each has around 350 frames and includes a model testing clothes of diverse appearances and textures.

\noindent - \texttt{BAIR} \cite{Ebert_2017}: a dataset of 43,008 videos with $256 \times 256$ size (42,880 for training, 128 for testing) collected by a Sawyer robotic arm pushing diverse objects, each has 30 frames.

\noindent - \texttt{MGif} \cite{Siarohin_2019_CVPR}: a dataset containing 1,000 videos (900 for training, 100 for testing) of different moving 2D cartoon animals, with each video as a \texttt{gif} file. 


\input{sub/plt/dfdc_recon} 

\vspace{-0.5cm}

\paragraph{Metrics}
\label{sec: metrics}
We utilize various metrics for extensively evaluating individual tasks across multiple aspects:


\noindent {\textbf{Reconstruction correctness (for video reconstruction):}}

\noindent - \texttt{L1}: the average $L1$ between the generated and real images, measuring the overall reconstruction correctness.

\noindent - \texttt{LPIPS} \cite{Zhang_2018_CVPR}: the perceptual distance between the generated and real images, which correlates better with human observations compared to \texttt{L1}. 

\noindent - \texttt{PSNR}: the mean squared error (MSE) between the generated and real images indicating reconstruction quality. 

\noindent - \texttt{MS-/SSIM}: \texttt{SSIM} measures the structural similarity between the generated and real images which is a more robust indicator than \texttt{PSNR} \cite{Wang_2021_CVPR}. \texttt{MS-SSIM} is a multi-scale variant of \texttt{SSIM} working on multiple scales of the images and has been shown to correlate well with human perception.

\noindent {\textbf{Video quality}} - \texttt{FID} (Fr\'echet Inception Distance): the distance between the distributions of the generated and real images. We use \texttt{PyTorch}'s official implementation \cite{Seitzer_2020_FID}, where the distribution distances are computed based on features extracted from the pre-trained \texttt{InceptionV3} network. For video reconstruction, the distance is measured with respect to the driving images, for image animation, it is measured with respect to the source.

\vspace{-0.02cm}
\noindent {\textbf{Semantic consistency}} - \texttt{AKD} (Average Key-point Distance): the average distance between keypoints detected from the generated and real images. For human body datasets, we use a pre-trained key-point detector~\cite{Cao_2017_CVPR,Siarohin_2019_NeurIPS}. For face datasets, we use the facial landmark detector~\cite{Bulat_2017_ICCV}.

\vspace{-0.02cm}
\noindent {\textbf{Animation identity preservation (for image animation)}} - \texttt{CSIM}: evaluates the quality of identity preservation. Specifically, it measures the cosine similarity between identity embedding vectors of the generated and real images. For human face datasets, the embeddings are extracted from a pre-trained face identity recognition model \cite{Deng_2019_CVPR,Ha_2020_AAAI,Wang_2021_CVPR}. For human body datasets, we use the identity embeddings from the pre-trained \texttt{PyTorch ReID} model \cite{Zheng_2019_CVPR,Siarohin_2019_NeurIPS}. 

\input{sub/plt/anim}  

\vspace{-0.0cm}
\subsection{Benchmark Results}
\label{sec: comparisons}







\vspace{-0.15cm}
Following previous work~\cite{Siarohin_2019_NeurIPS,Siarohin_2021_CVPR,Wang_2021_CVPR}, we adopt U-Net~\cite{Ronneberger2015UNetCN} as the feature extraction backbone in step \hyperlink{step-1}{1)}, and Johnson architecture~\cite{Johnson_2016_ECCV} as the image generator in step \hyperlink{step-3}{3)}. 
We train \texttt{DiME} following the strategy of \cite{Siarohin_2019_NeurIPS,Siarohin_2021_CVPR}, and re-train the baseline models \cite{Siarohin_2019_NeurIPS,Siarohin_2021_CVPR} using the same strategy. Throughout all experiments, we randomly select three reference images of the source object for \texttt{DiME}. For a \textit{fairer} comparison, we also report the best scores of using the \emph{same} reference images for one-shot baseline models \texttt{FOMM} \cite{Siarohin_2019_NeurIPS} and Articulated Animation (\texttt{AA}) \cite{Siarohin_2021_CVPR}. 
(in parentheses of Tab.~\ref{tab: all}), which could be viewed as the few-shot variant of baseline models. 

\vspace{-0.4cm}

\paragraph{Video Reconstruction}
\label{sec: reconstruction}

We first compare \texttt{DiME} with the state-of-the-arts for video reconstruction. Quantitative results are listed in Tab.~\ref{tab: all} (note that \texttt{face-vid2vid} \cite{Wang_2021_CVPR} is not applicable to datasets with objects other than talking heads), qualitative comparisons for \texttt{DFDC} are shown in Fig.~\ref{fig: dfdc_recon}. Clearly, \texttt{DiME} performs the best regarding \emph{every} metric across \emph{all} nine datasets, containing human faces, bodies, robots and cartoon animals. For the straightforward \texttt{L1} reconstruction error, our approach outperforms all state-of-the-art methods by nearly 30\% averagely. Using \texttt{LPIPS}, which was shown to correlate better with human observations \cite{Zhang_2018_CVPR}, our approach improves the state-of-the-art by approximately 30\% across all datasets. 

Notably, despite using additional reference images for one-shot baseline models \texttt{FOMM} \cite{Siarohin_2019_NeurIPS} and Articulated Animation (\texttt{AA}) \cite{Siarohin_2021_CVPR}, and picking the best results (in parentheses of Tab.~\ref{tab: all}) indeed help improve the overall performance of \texttt{FOMM} \cite{Siarohin_2019_NeurIPS} and \texttt{AA} \cite{Siarohin_2021_CVPR}, \texttt{DiME} still achieves the best scores.

\vspace{-0.3cm}
 
\paragraph{Image Animation}
\label{sec: animation}
We also report both quantitative (Tab.~\ref{tab: all}) and qualitative comparisons (Fig.~\ref{fig: anim}) for the image animation task. For each dataset, we randomly generated 200 pairs of videos containing difference identities as source-driving pairs. On all nine datasets, \texttt{DiME} outperforms the state-of-the-art. Particularly, it achieves clear improvements in \texttt{CSIM}, which reflects how well the identity is preserved in the generated video compared to the source image. This result could be observed qualitatively in Fig.~\ref{fig: anim}. In fact, we could observe that both \texttt{FOMM} \cite{Siarohin_2019_NeurIPS} and \texttt{AA} \cite{Siarohin_2021_CVPR} find it difficult to transfer the motion in the driving frame, which causes clear identity changes in the generated images. In contrast, \texttt{DiME} is able to transfer the motion and preserve the source identity better. 

Similar to video reconstruction (Sec.~\ref{sec: reconstruction}), additional reference images only marginally improve the animation performances of baseline models (\texttt{FOMM}, \texttt{AA}), while \texttt{DiME} outperforms all the compared models on all datasets. 


\input{sub/plt/vox-mgif-generality}

\begin{table}[!t] 
\centering
\renewcommand{\arraystretch}{1.2}
\resizebox{0.85\linewidth}{!}{
\begin{tabular}{cccccccc}
\specialrule{.15em}{.05em}{.05em}

{\textbf{\textit{Dataset}}} & {\textbf{\textit{Method}}} & \multicolumn{1}{c}{\multirow{1}{*}{\texttt{L1} ($\downarrow$)}} & \multicolumn{1}{c}{\multirow{1}{*}{\texttt{LPIPS} ($\downarrow$)}} & \multicolumn{1}{c}{\multirow{1}{*}{\texttt{FID} ($\downarrow$)}} & \multicolumn{1}{c}{\multirow{1}{*}{\texttt{MS-/SSIM} ($\uparrow$)}} & \multicolumn{1}{c}{\multirow{1}{*}{\texttt{PSNR} ($\uparrow$)}} & \multicolumn{1}{c}{\multirow{1}{*}{\texttt{AKD} ($\downarrow$)}} \\

\specialrule{.08em}{.05em}{.05em}

\multicolumn{1}{c}{\multirow{3}{*}{\shortstack{\texttt{Nemo}\\(124)}}} &  \texttt{FOMM} \cite{Siarohin_2019_NeurIPS} & 0.0191  & 0.047 & 37.57 & 0.94/0.91 & 29.93 & 1.36    \\ 

& \texttt{AA} \cite{Siarohin_2021_CVPR} & 0.0170 & 0.045 & 33.95 & 0.95/0.92 & 30.38 & 1.24 \\  


& \texttt{DiME} & {\bf{0.0086}} & {\bf{0.023}} & {\bf{29.45}} & {\bf{0.99}}/{\bf{0.97}} & {\bf{36.80}} & {\bf{0.98}}   \\

\hline
 
\multicolumn{1}{c}{\multirow{3}{*}{\shortstack{\texttt{MGif}\\(100)}}} & \texttt{FOMM} \cite{Siarohin_2019_NeurIPS} & 0.086 & 0.256 & 222.99 & 0.76/0.63 & 15.38 & N/A  \\ 

& \texttt{AA} \cite{Siarohin_2021_CVPR} & 0.102 & 0.345 & 269.44 & 0.57/0.73 & 14.82 & N/A \\ 

& \texttt{DiME} & {\bf{0.045}} & {\bf{0.176}} & {\bf{158.96}} & {\bf{0.81/0.86}} & {\bf{18.70}} & N/A \\

\specialrule{.15em}{.05em}{.05em} 
\end{tabular} 
}
\vspace{-0.25cm}
\caption{Model generality results for \texttt{FOMM} \cite{Siarohin_2019_NeurIPS}, Articulated Animation (\texttt{AA}) \cite{Siarohin_2021_CVPR} and \texttt{DiME}, computed from models trained on \texttt{VoxCeleb}, but tested on \texttt{Nemo} (face) and \texttt{MGif} (cartoons). 
}
\label{tab: gen_all}
\end{table}

\vspace{-0.3cm}

\paragraph{Zero-shot Motion Transfer}
\label{sec: generality}
We further explore the generality of \texttt{DiME}, and compare its zero-shot motion transfer performance with the state-of-the-art approaches. Specifically, we are interested in the scenario where models are trained on one dataset but tested on an another dataset with different object category that has \textit{not} been seen during training (Fig.~\ref{fig: vox_mgif_gen}). We have already compared the reconstruction results for models that were trained on \texttt{VoxCeleb} (human faces) dataset~(Tab.~\ref{tab: all}), here, we use the same models trained from \texttt{VoxCeleb} but test them on \texttt{Nemo} (haman faces) and \texttt{MGif} (cartoons) dataset (Tab.~\ref{tab: gen_all}). As illustrated in Tab.~\ref{tab: gen_all}, \texttt{DiME} achieves significant improvements across all metrics over the state-of-the-art models. Note that on \texttt{Nemo} dataset, \texttt{DiME} that trained on \texttt{VoxCeleb} dataset even outperforms the state-of-the-art models that are specifically trained on \texttt{Nemo} dataset (see Tab.~\ref{tab: all}). As shown in Fig.~\ref{fig: vox_mgif_gen}, when tested on doma ins different from \texttt{VoxCeleb} (human faces), both \texttt{FOMM} \cite{Siarohin_2019_NeurIPS} and \texttt{AA} \cite{Siarohin_2021_CVPR} struggle to preserve the source identities of cartoon animals.

\vspace{-0.0cm}
\subsection{Ablation Studies}
\label{sec: ablation}


\begin{table}[!t] 
\centering
\renewcommand{\arraystretch}{1.2}
\resizebox{\linewidth}{!}{
\begin{tabular}{ccccccccc}
\specialrule{.15em}{.05em}{.05em}
 
\multicolumn{1}{c}{\multirow{2}{*}{\textbf{\textit{Model}}}} & \multicolumn{6}{c}{\multirow{1}{*}{\textbf{\textit{Reconstruction}}}} & \multicolumn{2}{c}{\multirow{1}{*}{\textbf{\textit{Animation}}}} \\

\cmidrule(lr){2-7}
\cmidrule(lr){8-9}

 & \multicolumn{1}{c}{\multirow{1}{*}{\texttt{L1} ($\downarrow$)}} & \multicolumn{1}{c}{\multirow{1}{*}{\texttt{LPIPS} ($\downarrow$)}} & \multicolumn{1}{c}{\multirow{1}{*}{\texttt{FID} ($\downarrow$)}} & \multicolumn{1}{c}{\multirow{1}{*}{\texttt{MS-/SSIM} ($\uparrow$)}} & \multicolumn{1}{c}{\multirow{1}{*}{\texttt{PSNR} ($\uparrow$)}} & \multicolumn{1}{c}{\multirow{1}{*}{\texttt{AKD} ($\downarrow$)}} & \multicolumn{1}{c}{\multirow{1}{*}{\texttt{FID} ($\downarrow$)}} & \multicolumn{1}{c}{\multirow{1}{*}{\texttt{CSIM} ($\uparrow$)}}\\

\specialrule{.08em}{.05em}{.05em}

\textit{\texttt{DiME} (Full model)} & {\bf{0.018}}  & {\bf{0.0061}} & {\bf{33.22}} & {\bf{0.95}}/{\bf{0.91}} & 27.99 & {\bf{1.28}} & {\bf{64.89}} & {\bf{0.92}} \\  

\texttt{w/o refine} & 0.022  & 0.082   & 42.47 & 0.93/0.87 & {\bf{28.60}} & 1.53 & 71.88 & 0.88  \\ 

\texttt{w/o ID} & 0.024   & 0.085   & 41.89 & 0.92/0.85 & 27.88 & 1.29 & 75.35 & 0.90  \\ 

\texttt{one-shot} & 0.032   & 0.108   & 54.57 & 0.88/0.84 & 24.24 & 1.78 & 80.34 & 0.89  \\

\specialrule{.15em}{.05em}{.05em} 
\end{tabular}
}

\vspace{-0.25cm}
\caption{Ablations of \texttt{DiME} on \texttt{DFDC} (1807 testing samples for reconstruction, 200 testing pairs for animation), with mean results reported and best ones bolded. 
}

\label{tab: ablations}
\end{table}



\vspace{-0.15cm}

\paragraph{Motion Evolution Contributions}
\label{sec: component}
\input{sub/plt/dfdc_ablation}

In order to analyze how \texttt{DiME} contributes to the final performance, we design the three variants of the proposed \texttt{DiME} model: (1) \texttt{w/o refine}: \texttt{DiME} without the motion deformation refinement~(Sec.~\ref{sec: refine}), i.e., using Eq.~(\ref{eq: initial_dense}) as the final estimation for the motion deformation field; (2) \texttt{w/o ID}: \texttt{DiME} without being conditioned on the source identity~(Sec.~\ref{sec: appearance}), i.e., using only $\mathbf{F}_{\text{motion}}$ in Eq.~(\ref{eq: multi_view}) for image generation; (3) \texttt{one-shot}: \texttt{DiME} without additional reference image, i.e., the one-shot version of \texttt{DiME}, with only the source image as the input from the source domain. As demonstrated in Tab.~\ref{tab: ablations}, the full \texttt{DiME} model performs the best among all model variants, and ignoring any of the proposed designs in \texttt{DiME} would harm the model's overall motion transfer performance. 
We also provide qualitative comparisons of the above three model variants with the full \texttt{DiME} model~(Fig.~\ref{fig: dfdc_ablation}). Without motion refinement, the predicted dense motion deformation fields struggle to capture \textit{large} motion changes between the source and driving domains (e.g., headpose and mouth shape in 2$^{\text{nd}}$ row). Without conditioning on the source identity, we observe \textit{unrealistic} patterns in the missing regions (e.g., facial details in 1$^{\text{st}}$ row). With only one source viewpoint, it is challenging to realistically deform the details in source image to the driving domain (e.g., facial expressions in 3$^{\text{rd}}$ row).

\begin{table}[!t] 
\centering
\renewcommand{\arraystretch}{1.2}
\resizebox{0.85\linewidth}{!}{
\begin{tabular}{ccccccccc}
\specialrule{.15em}{.05em}{.05em}

\multicolumn{1}{l}{\multirow{2}{*}{{$\mathbf{N}_{\text{ref}}$}}} & \multicolumn{6}{c}{\multirow{1}{*}{\textbf{\textit{Reconstruction}}}} & \multicolumn{2}{c}{\multirow{1}{*}{\textbf{\textit{Animation}}}} \\

\cmidrule(lr){2-7}
\cmidrule(lr){8-9}

 & \multicolumn{1}{c}{\multirow{1}{*}{\texttt{L1} ($\downarrow$)}} & \multicolumn{1}{c}{\multirow{1}{*}{\texttt{LPIPS} ($\downarrow$)}} & \multicolumn{1}{c}{\multirow{1}{*}{\texttt{FID} ($\downarrow$)}} & \multicolumn{1}{c}{\multirow{1}{*}{\texttt{MS-/SSIM} ($\uparrow$)}} & \multicolumn{1}{c}{\multirow{1}{*}{\texttt{PSNR} ($\uparrow$)}} & \multicolumn{1}{c}{\multirow{1}{*}{\texttt{AKD} ($\downarrow$)}} & \multicolumn{1}{c}{\multirow{1}{*}{\texttt{FID} ($\downarrow$)}} & \multicolumn{1}{c}{\multirow{1}{*}{\texttt{CSIM} ($\uparrow$)}}\\

\specialrule{.08em}{.05em}{.05em}

\textit{1} & 0.024   & 0.088   & 42.20 & 0.90/0.87 & 27.23 & 1.33 & 68.78 & 0.89 \\ 

\textit{2} & 0.021   & 0.070   & 35.34 & 0.93/0.89 & 27.35 & 1.34 & 67.58 & 0.90 \\ 

\textit{3} & 0.018 & 0.061  & 33.22 & 0.95/0.91 & 27.99 & 1.28 & 64.89 & 0.92 \\ 

\textit{4} & 0.017   & 0.058   & 31.89 & 0.95/0.92 & 28.56 & 1.24 & 64.82 & 0.92 \\ 

\textit{5} & 0.017   & 0.051   & 30.69 & 0.96/0.92 & 28.83 & 1.21 & 65.15 & 0.92 \\ 

\textit{6} &   0.016  &  0.050   &  31.17 &  0.96/0.92  &  29.04 & 1.19  & 64.86 & 0.92   \\

\specialrule{.15em}{.05em}{.05em}

\end{tabular}  
}

\vspace{-0.25cm}
\caption{Reconstruction \& animation results for different number of additional reference image ($N_{\text{ref}}$) (\texttt{DFDC}, 1807 testing samples for reconstruction task, 200 testing pairs for animation), with mean results reported and best ones bolded. 
}

\label{tab: num_refs_all}
\end{table}

\vspace{-0.4cm}
\paragraph{Number of Reference Images}
\label{sec: num_refs}

In essence, optimized in an ODE system, the number of reference images ($N_{\text{ref}}$) \texttt{DiME} uses in inference is \textit{not} restricted by that in training. 
Here we explore \texttt{DiME}'s performance with varying $N_{\text{ref}}$ during inference. As shown in Tab.~\ref{tab: num_refs_all}, for reconstruction, increasing the number of reference images generally helps improve the model's performance. On the other hand, for animation, increasing the number of reference images, which lead to higher memory cost, did not necessarily boost performance. Based on our experiments, three reference images could achieve reasonably good performance for both the reconstruction and animation tasks.

\vspace{-0.1cm}

\section{Conclusions}
\label{sec: con}
\vspace{-0.15cm}
We proposed \texttt{DiME}, an end-to-end, unsupervised motion transfer framework designed to handle large motion changes. \texttt{DiME} formulates motion transfer as regularized optimization with an ODE system, where the integration of ODEs differentially refine the motion deformations, and intrinsically results in its flexible one/few-shot setup. 
We further condition the motion warping on features from the source identity, resulting in more realistic generations in the missing regions. Experiments across nine datasets with varying objects consistently demonstrate \texttt{DiME}'s superiority over the state-of-the-arts. Zero-shot testing on model's generality further shows that \texttt{DiME} performs well and stably on novel objects that it has not seen during training.



{\small
\bibliographystyle{ieee_fullname}
\bibliography{references}

\begin{thebibliography}{10}\itemsep=-1pt

\bibitem{benamou2000omt}
Jean-David Benamou and Yann Brenier.
\newblock A computational fluid mechanics solution to the monge-kantorovich mass transfer problem.
\newblock {\em Numerische Mathematik}, 84, 2000.

\bibitem{Bhatnagar_2019_ICCV}
Bharat~Lal Bhatnagar, Garvita Tiwari, Christian Theobalt, and Gerard Pons-Moll.
\newblock Multi-garment net: Learning to dress 3d people from images.
\newblock In {\em {IEEE} International Conference on Computer Vision ({ICCV})}, 2019.

\bibitem{Blanz_1999_ACCGIT}
Volker Blanz and Thomas Vetter.
\newblock A morphable model for the synthesis of 3d faces.
\newblock In {\em Annual Conference on Computer Graphics and Interactive Techniques}, 1999.

\bibitem{Bulat_2017_ICCV}
Adrian Bulat and Georgios Tzimiropoulos.
\newblock How far are we from solving the 2d \& 3d face alignment problem? (and a dataset of 230,000 3d facial landmarks).
\newblock In {\em International Conference on Computer Vision (ICCV)}, 2017.

\bibitem{Buttazzo2009AnOP}
Giorgio~C. Buttazzo, Chlo{\'e} Jimenez, and {\'E}douard Oudet.
\newblock An optimization problem for mass transportation with congested dynamics.
\newblock {\em SIAM J. Control. Optim.}, 48:1961--1976, 2009.

\bibitem{Cao_2014_TG}
Chen Cao, Qiming Hou, and Kun Zhou.
\newblock Displaced dynamic expression regression for real-time facial tracking and animation.
\newblock {\em ACM Trans. Graph.}, 2014.

\bibitem{Cao_2017_CVPR}
Zhe Cao, Tomas Simon, Shih-En Wei, and Yaser Sheikh.
\newblock Realtime multi-person 2d pose estimation using part affinity fields.
\newblock In {\em IEEE/CVF Conference on Computer Vision and Pattern Recognition (CVPR)}, 2017.

\bibitem{Chan_2019_ICCV}
Caroline Chan, Shiry Ginosar, Tinghui Zhou, and Alexei~A Efros.
\newblock Everybody dance now.
\newblock In {\em IEEE International Conference on Computer Vision (ICCV)}, 2019.

\bibitem{Chen_2018_Neurips}
Ricky T.~Q. Chen, Yulia Rubanova, Jesse Bettencourt, and David Duvenaud.
\newblock Neural ordinary differential equations.
\newblock In {\em Advances in Neural Information Processing Systems (NeurIPS)}, 2018.

\bibitem{Chen2016OnTR}
Yongxin Chen, Tryphon~T. Georgiou, and Michele Pavon.
\newblock On the relation between optimal transport and schr{\"o}dinger bridges: A stochastic control viewpoint.
\newblock {\em Journal of Optimization Theory and Applications}, 169:671--691, 2016.

\bibitem{NIPS2013_af21d0c9}
Marco Cuturi.
\newblock Sinkhorn distances: Lightspeed computation of optimal transport.
\newblock In C.J. Burges, L. Bottou, M. Welling, Z. Ghahramani, and K.Q. Weinberger, editors, {\em Advances in Neural Information Processing Systems}, volume~26. Curran Associates, Inc., 2013.

\bibitem{Deng_2019_CVPR}
Jiankang Deng, Jia Guo, Xue Niannan, and Stefanos Zafeiriou.
\newblock Arc{F}ace: Additive angular margin loss for deep face recognition.
\newblock In {\em IEEE/CVF Conference on Computer Vision and Pattern Recognition (CVPR)}, 2019.

\bibitem{Hamdi_2012_ECCV}
Hamdi Dibeklio{\u{g}}lu, Albert~Ali Salah, and Theo Gevers.
\newblock Are you really smiling at me? spontaneous versus posed enjoyment smiles.
\newblock In {\em European Conference on Computer Vision (ECCV)}, 2012.

\bibitem{Hamdi_2015_Tran_MM}
Hamdi Dibeklio{\u{g}}lu, Albert~Ali Salah, and Theo Gevers.
\newblock Recognition of genuine smiles.
\newblock {\em IEEE Transactions on Multimedia}, 17(3):279--294, 2015.

\bibitem{DFDC_2020}
Brian Dolhansky, Joanna Bitton, Ben Pflaum, Jikuo Lu, Russ Howes, Menglin Wang, and Cristian~Canton Ferrer.
\newblock The deepfake detection challenge dataset.
\newblock {\em arXiv Preprint}, 2020.

\bibitem{Ebert_2017}
Frederik Ebert, Chelsea Finn, Alex~X. Lee, and Sergey Levine.
\newblock Self-supervised visual planning with temporal skip connections.
\newblock In {\em Conference on Robot Learning (CoRL)}, 2017.

\bibitem{Feydy2017OptimalTF}
Jean Feydy, Benjamin Charlier, François-Xavier Vialard, and Gabriel Peyr{\'e}.
\newblock Optimal transport for diffeomorphic registration.
\newblock In {\em MICCAI}, 2017.

\bibitem{Fitschen2016TransportBR}
Jan~Henrik Fitschen, Friederike Laus, and Gabriele Steidl.
\newblock Transport between rgb images motivated by dynamic optimal transport.
\newblock {\em Journal of Mathematical Imaging and Vision}, 56:409--429, 2016.

\bibitem{Geng_2019_CVPR}
Zhenglin Geng, Chen Cao, and Sergey Tulyakov.
\newblock 3d guided fine-grained face manipulation.
\newblock In {\em IEEE/CVF Conference on Computer Vision and Pattern Recognition (CVPR)}, 2019.

\bibitem{Ha_2020_AAAI}
Sungjoo Ha, Martin Kersner, Beomsu Kim, Seokjun Seo, and Dongyoung Kim.
\newblock Mario{NET}te: Few-shot face reenactment preserving identity of unseen targets.
\newblock In {\em Proceedings of the AAAI Conference on Artificial Intelligence (AAAI)}, 2020.

\bibitem{Haker2001MassPM}
Steven Haker, Allen~R. Tannenbaum, and Ron Kikinis.
\newblock Mass preserving mappings and image registration.
\newblock In {\em MICCAI}, 2001.

\bibitem{He_2019_CVPR}
Xiangyu He, Zitao Mo, Peisong Wang, Yang Liu, Mingyuan Yang, and Jian Cheng.
\newblock Ode-inspired network design for single image super-resolution.
\newblock In {\em IEEE/CVF Conference on Computer Vision and Pattern Recognition (CVPR)}, 2019.

\bibitem{Jakab_2018_Neurips}
Tomas Jakab, Ankush Gupta, Hakan Bilen, and Andrea Vedaldi.
\newblock Unsupervised learning of object landmarks through conditional image generation.
\newblock In {\em Advances in Neural Information Processing Systems (NeurIPS)}, 2018.

\bibitem{Johnson_2016_ECCV}
Justin Johnson, Alexandre Alahi, and Li Fei-Fei.
\newblock Perceptual losses for real-time style transfer and super-resolution.
\newblock In {\em European Conference of Computer vision (ECCV)}, 2016.

\bibitem{Kantorovitch1958OnTT}
L. Kantorovitch.
\newblock On the translocation of masses.
\newblock {\em Management Science}, 5:1--4, 1958.

\bibitem{Koundal2020OptimalMT}
Sunil Koundal, Rena Elkin, Saad Nadeem, Yuechuan Xue, Stefan Constantinou, Simon Sanggaard, Xiaodan Liu, Brittany Monte, Feng Xu, William E.~Van Nostrand, Maiken Nedergaard, Hedok Lee, Joanna~Marguerite Wardlaw, H{\'e}l{\`e}ne Benveniste, and Allen~R. Tannenbaum.
\newblock Optimal mass transport with lagrangian workflow reveals advective and diffusion driven solute transport in the glymphatic system.
\newblock {\em Scientific Reports}, 10, 2020.

\bibitem{Lee_2019_ICLR}
Jessica Lee, Deva Ramanan, and Rohit Girdhar.
\newblock {MetaPix: Few-Shot Video Retargeting}.
\newblock {\em ICLR}, 2020.

\bibitem{Li_2019_CVPR}
Yining Li, Chen Huang, and Chen~Change Loy.
\newblock Dense intrinsic appearance flow for human pose transfer.
\newblock In {\em IEEE/CVF Conference on Computer Vision and Pattern Recognition (CVPR)}, 2019.

\bibitem{Liu_2022_CVPR}
Peirong Liu, Yueh Lee, Stephen Aylward, and Marc Niethammer.
\newblock Deep decomposition for stochastic normal-abnormal transport.
\newblock In {\em IEEE/CVF Conference on Computer Vision and Pattern Recognition (CVPR)}, 2022.

\bibitem{Liu_2021_CVPR}
Peirong Liu, Lin Tian, Yubo Zhang, Stephen Aylward, Yueh Lee, and Marc Niethammer.
\newblock Discovering hidden physics behind transport dynamics.
\newblock In {\em IEEE/CVF Conference on Computer Vision and Pattern Recognition (CVPR)}, 2021.

\bibitem{Liu_2019_ICCV}
Wen Liu, Zhixin Piao, Min Jie, Wenhan Luo, Lin Ma, and Shenghua Gao.
\newblock Liquid warping gan: A unified framework for human motion imitation, appearance transfer and novel view synthesis.
\newblock In {\em IEEE International Conference on Computer Vision (ICCV)}, 2019.

\bibitem{Loper_2015_SIGGRAPH}
Matthew Loper, Naureen Mahmood, Javier Romero, Gerard Pons-Moll, and Michael~J. Black.
\newblock {SMPL}: A skinned multi-person linear model.
\newblock {\em ACM Trans. Graphics (Proc. SIGGRAPH Asia)}, 2015.

\bibitem{Lu_2018_ICML}
Yiping Lu, Aoxiao Zhong, Quanzheng Li, and Bin Dong.
\newblock Beyond finite layer neural networks: Bridging deep architectures and numerical differential equations.
\newblock In {\em International Conference on Machine Learning (ICML)}, 2018.

\bibitem{Ma_2020_CVPR}
Qianli Ma, Jinlong Yang, Anurag Ranjan, Sergi Pujades, Gerard Pons-Moll, Siyu Tang, and Michael~J. Black.
\newblock Learning to dress 3d people in generative clothing.
\newblock In {\em IEEE/CVF Conference on Computer Vision and Pattern Recognition (CVPR)}, 2020.

\bibitem{Mir_2020_CVPR}
Aymen Mir, Thiemo Alldieck, and Gerard Pons-Moll.
\newblock Learning to transfer texture from clothing images to 3d humans.
\newblock In {\em IEEE/CVF Conference on Computer Vision and Pattern Recognition (CVPR)}. {IEEE}, 2020.

\bibitem{Nagrani_2017}
A. Nagrani, J.~S. Chung, and A. Zisserman.
\newblock Voxceleb: a large-scale speaker identification dataset.
\newblock In {\em INTERSPEECH}, 2017.

\bibitem{Oquab_2020_arxiv}
Maxime Oquab, Pierre Stock, Oran Gafni, Daniel Haziza, Tao Xu, Peizhao Zhang, Onur Celebi, Yana Hasson, Patrick Labatut, Bobo Bose-Kolanu, Thibault Peyronel, and Camille Couprie.
\newblock Low bandwidth video-chat compression using deep generative models.
\newblock {\em arXiv Preprint}, 2020.

\bibitem{Paoletti_2020_TGRS}
Mercedes~E. Paoletti, Juan~Mario Haut, Javier Plaza, and Antonio Plaza.
\newblock Neural ordinary differential equations for hyperspectral image classification.
\newblock {\em IEEE Transactions on Geoscience and Remote Sensing}, 2020.

\bibitem{Patel_2020_CVPR}
Chaitanya Patel, Zhouyingcheng Liao, and Gerard Pons-Moll.
\newblock Tailornet: Predicting clothing in 3d as a function of human pose, shape and garment style.
\newblock In {\em IEEE/CVF Conference on Computer Vision and Pattern Recognition (CVPR)}. {IEEE}, 2020.

\bibitem{Qian_2019_ICCV}
Shengju Qian, Kwan{-}Yee Lin, Wayne Wu, Yangxiaokang Liu, Quan Wang, Fumin Shen, Chen Qian, and Ran He.
\newblock Make a face: Towards arbitrary high fidelity face manipulation.
\newblock In {\em 2019 {IEEE/CVF} International Conference on Computer Vision, {ICCV}}, 2019.

\bibitem{Ren_2021_CVPR}
Jian Ren, Menglei Chai, Oliver~J. Woodford, Kyle Olszewski, and Sergey Tulyakov.
\newblock Flow guided transformable bottleneck networks for motion retargeting.
\newblock In {\em IEEE/CVF Conference on Computer Vision and Pattern Recognition (CVPR)}, 2021.

\bibitem{Ronneberger2015UNetCN}
Olaf Ronneberger, Philipp Fischer, and Thomas Brox.
\newblock U-net: Convolutional networks for biomedical image segmentation.
\newblock {\em ArXiv}, abs/1505.04597, 2015.

\bibitem{Ruthotto_2018_JMIV}
Lars Ruthotto and Eldad Haber.
\newblock Deep neural networks motivated by partial differential equations.
\newblock {\em Journal of Mathematical Imaging and Vision}, 2018.

\bibitem{Seitzer_2020_FID}
Maximilian Seitzer.
\newblock {pytorch-fid: FID Score for PyTorch}.
\newblock \url{https://github.com/mseitzer/pytorch-fid}, August 2020.
\newblock Version 0.1.1.

\bibitem{Shalev2020ImageAW}
Yoav Shalev and Lior Wolf.
\newblock Image animation with perturbed masks.
\newblock {\em 2022 IEEE/CVF Conference on Computer Vision and Pattern Recognition (CVPR)}, pages 3637--3646, 2020.

\bibitem{Siarohin2018AnimatingAO}
Aliaksandr Siarohin, St{\'e}phane Lathuili{\`e}re, S. Tulyakov, Elisa Ricci, and N. Sebe.
\newblock Animating arbitrary objects via deep motion transfer.
\newblock {\em 2019 IEEE/CVF Conference on Computer Vision and Pattern Recognition (CVPR)}, pages 2372--2381, 2018.

\bibitem{Siarohin_2019_CVPR}
Aliaksandr Siarohin, Stéphane Lathuilière, Sergey Tulyakov, Elisa Ricci, and Nicu Sebe.
\newblock Animating arbitrary objects via deep motion transfer.
\newblock In {\em IEEE/CVF Conference on Computer Vision and Pattern Recognition (CVPR)}, 2019.

\bibitem{Siarohin_2019_NeurIPS}
Aliaksandr Siarohin, Stéphane Lathuilière, Sergey Tulyakov, Elisa Ricci, and Nicu Sebe.
\newblock First order motion model for image animation.
\newblock In {\em Conference on Neural Information Processing Systems (NeurIPS)}, 2019.

\bibitem{Siarohin_2021_CVPR}
Aliaksandr Siarohin, Oliver Woodford, Jian Ren, Menglei Chai, and Sergey Tulyakov.
\newblock Motion representations for articulated animation.
\newblock In {\em IEEE/CVF Conference on Computer Vision and Pattern Recognition (CVPR)}, 2021.

\bibitem{Tao2022StructureAwareMT}
Jiale Tao, Biao Wang, Borun Xu, Tiezheng Ge, Yuning Jiang, Wen Li, and Lixin Duan.
\newblock Structure-aware motion transfer with deformable anchor model.
\newblock {\em 2022 IEEE/CVF Conference on Computer Vision and Pattern Recognition (CVPR)}, pages 3627--3636, 2022.

\bibitem{Thies_2016_CVPR}
J. Thies, M. Zollh{\"o}fer, M. Stamminger, C. Theobalt, and M. Nie{\ss}ner.
\newblock {Face2Face: Real-time Face Capture and Reenactment of RGB Videos}.
\newblock In {\em IEEE/CVF Conference on Computer Vision and Pattern Recognition (CVPR)}, 2016.

\bibitem{Torres2021ASO}
Luis~Caicedo Torres, Luiz~Manella Pereira, and Mostafa Amini.
\newblock A survey on optimal transport for machine learning: Theory and applications.
\newblock {\em ArXiv}, abs/2106.01963, 2021.

\bibitem{AIST_2019}
Shuhei Tsuchida, Satoru Fukayama, Masahiro Hamasaki, and Masataka Goto.
\newblock {AIST} dance video database: Multi-genre, multi-dancer, and multi-camera database for dance information processing.
\newblock In {\em International Society for Music Information Retrieval Conference (ISMIR)}, 2019.

\bibitem{Rafael_2019_arxiv}
Rafael Valle, Fitsum Reda, Mohammad Shoeybi, Patrick Legresley, Andrew Tao, and Bryan Catanzaro.
\newblock Neural odes for image segmentation with level sets.
\newblock {\em arXiv Preprint}, 2019.

\bibitem{Wang_2019_Neurips}
Ting-Chun Wang, Ming-Yu Liu, Andrew Tao, Guilin Liu, Jan Kautz, and Bryan Catanzaro.
\newblock Few-shot video-to-video synthesis.
\newblock In {\em Advances in Neural Information Processing Systems (NeurIPS)}, 2019.

\bibitem{Wang_2018_Neurips}
Ting-Chun Wang, Ming-Yu Liu, Jun-Yan Zhu, Guilin Liu, Andrew Tao, Jan Kautz, and Bryan Catanzaro.
\newblock Video-to-video synthesis.
\newblock In {\em Advances in Neural Information Processing Systems (NeurIPS)}, 2018.

\bibitem{Wang_2021_CVPR}
Ting-Chun Wang, Arun Mallya, and Ming-Yu Liu.
\newblock One-shot free-view neural talking-head synthesis for video conferencing.
\newblock In {\em IEEE/CVF Conference on Computer Vision and Pattern Recognition (CVPR)}, 2021.

\bibitem{Wang_2018_CVPR}
W. Wang, X. Alameda-Pineda, D. Xu, P. Fua, E. Ricci, and N. Sebe.
\newblock Every smile is unique: Landmark-guided diverse smile generation.
\newblock In {\em IEEE/CVF Conference on Computer Vision and Pattern Recognition (CVPR)}, 2018.

\bibitem{Wang2022LatentIA}
Yaohui Wang, Di Yang, François Br{\'e}mond, and Antitza Dantcheva.
\newblock Latent image animator: Learning to animate images via latent space navigation.
\newblock {\em ArXiv}, abs/2203.09043, 2022.

\bibitem{Wiles2018X2FaceAN}
Olivia Wiles, A.~Sophia Koepke, and Andrew Zisserman.
\newblock X2face: A network for controlling face generation by using images, audio, and pose codes.
\newblock In {\em European Conference on Computer Vision}, 2018.

\bibitem{Wiles_2018_ECCV}
Olivia Wiles, A.~Sophia Koepke, and Andrew Zisserman.
\newblock X2face: A network for controlling face generation using images, audio, and pose codes.
\newblock In {\em European Conference on Computer Vision (ECCV)}, 2018.

\bibitem{Ali_2019_arxiv}
Ali~Pour Yazdanpanah and Simon K.~Warfield Onur~Afacan.
\newblock Ode-based deep network for mri reconstruction.
\newblock {\em arXiv Preprint}, 2019.

\bibitem{Zablotskaia_2019}
Polina Zablotskaia, Aliaksandr Siarohin, Bo Zhao, and Leonid Sigal.
\newblock Dwnet: Dense warp-based network for pose-guided human video generation.
\newblock In {\em British Machine Vision Conference (BMVC)}, 2019.

\bibitem{Zakharov_2020_ECCV}
Egor Zakharov, Aleksei Ivakhnenko, Aliaksandra Shysheya, and Victor Lempitsky.
\newblock Fast bi-layer neural synthesis of one-shot realistic head avatars.
\newblock In {\em European Conference of Computer vision (ECCV)}, 2020.

\bibitem{Zakharov_2019_CVPR}
Egor Zakharov, Aliaksandra Shysheya, Egor Burkov, and Victor Lempitsky.
\newblock Few-shot adversarial learning of realistic neural talking head models.
\newblock In {\em IEEE/CVF Conference on Computer Vision and Pattern Recognition (CVPR)}, 2019.

\bibitem{Zhang2021ARO}
Jingyi Zhang, Wenxuan Zhong, and Ping Ma.
\newblock A review on modern computational optimal transport methods with applications in biomedical research.
\newblock {\em ArXiv}, abs/2008.02995, 2021.

\bibitem{Zhang_2018_CVPR}
Richard Zhang, Phillip Isola, Alexei~A Efros, Eli Shechtman, and Oliver Wang.
\newblock The unreasonable effectiveness of deep features as a perceptual metric.
\newblock In {\em IEEE/CVF Conference on Computer Vision and Pattern Recognition (CVPR)}, 2018.

\bibitem{Zhang_2019_Neurips}
Tianjun Zhang, Zhewei Yao, Amir Gholami, Joseph~E Gonzalez, Kurt Keutzer, Michael~W Mahoney, and George Biros.
\newblock Anodev2: A coupled neural ode framework.
\newblock In H. Wallach, H. Larochelle, A. Beygelzimer, F. d\textquotesingle Alch\'{e}-Buc, E. Fox, and R. Garnett, editors, {\em Advances in Neural Information Processing Systems (NeurIPS)}, 2019.

\bibitem{Zhao2022ThinPlateSM}
Jian Zhao and Hui Zhang.
\newblock Thin-plate spline motion model for image animation.
\newblock {\em 2022 IEEE/CVF Conference on Computer Vision and Pattern Recognition (CVPR)}, pages 3647--3656, 2022.

\bibitem{Zheng_2019_CVPR}
Zhedong Zheng, Xiaodong Yang, Zhiding Yu, Liang Zheng, Yi Yang, and Jan Kautz.
\newblock Joint discriminative and generative learning for person re-identification.
\newblock In {\em IEEE/CVF Conference on Computer Vision and Pattern Recognition (CVPR)}, 2019.

\bibitem{Zhou_2016_ECCV}
Tinghui Zhou, Shubham Tulsiani, Weilun Sun, Jitendra Malik, and Alexei~A Efros.
\newblock View synthesis by appearance flow.
\newblock In {\em European Conference on Computer Vision (ECCV)}, 2016.

\bibitem{Zollhofer_2018_CGF}
M. Zollh\"ofer, J. Thies, P. Garrido, D. Bradley, T. Beeler, P. P\'erez, M. Stamminger, M. Nie\ss~ner, and C. Theobalt.
\newblock State of the art on monocular 3d face reconstruction, tracking, and applications.
\newblock {\em Computer Graphics Forum}, 2018.

\end{thebibliography}
}

\renewcommand\thefigure{\arabic{figure}}

\clearpage
\twocolumn[
\centering
\Large
\textbf{Differential Motion Evolution for Fine-Grained Motion Deformation in\\Unsupervised Image Animation (Appendix)}\\
\vspace{1.5em}
]

\setcounter{figure}{1}
\setcounter{equation}{1}

\renewcommand\thefigure{\thesection.\arabic{figure}}
\renewcommand\theequation{\thesection.\arabic{equation}}

\appendix

\section{Motion Deformation Estimation in an ODE System: Existence and Uniqueness}
\label{app_sec: ode_supp}

\paragraph{Initial condition.} 

As introduced in Sec.~\hyperlink{sec: keypoint}{3.2}, 
\texttt{DiME} first obtains coarse motion by extracting $K$ corresponding key-points from the source ($\mathbb{S}$) and reference images ($\mathbb{R}$) of the source object, as well as from the driving object ($\mathbb{D}$), respectively. The resulting motion deformation field,
\begin{equation}
\varphi_{\mathbb{S} \rightarrow \mathbb{D}}, \,\, \varphi_{\mathbb{R}^{(i)} \rightarrow \mathbb{D}},\quad (i = 1,\, \dots,\, N_{\text{ref}}),
\label{app-eq: initial_dense}
\end{equation}
is then viewed as the initial value of the motion estimation ODE system.

\paragraph{Motion refinement in an ODE system.}
As described in Sec.~\hyperlink{sec: refine}{3.2}, 
given the initial value of coarse motion deformation predictions in Eq.~(\ref{app-eq: initial_dense}), \texttt{DiME} gradually refines Eq.~(\ref{app-eq: initial_dense}) in a system of non-linear ODEs. I.e.,  ($\epsilon \in [0,\, 1]$, $i = 1,\, \dots,\, N_{\text{ref}}$)
\begin{equation}
\begin{cases}
    \frac{d \phi_{\mathbb{S} \rightarrow \mathbb{D}}(\epsilon)}{d \epsilon} = f_0 (\phi_{\mathbb{S} \rightarrow \mathbb{D}}(\epsilon),\, \epsilon)\,, \\ \vspace{-0.1cm}
    \frac{d \phi_{\mathbb{R}^{(i)} \rightarrow \mathbb{D}}(\epsilon)}{d \epsilon} = f_i (\phi_{\mathbb{R}^{(i)} \rightarrow \mathbb{D}}(\epsilon),\, \epsilon)\,,
\label{app-eq: ode_main}
\end{cases}
\end{equation}
\vspace{-0.1cm}
\begin{equation}
s.t.~
    \begin{cases}
      \phi_{\mathbb{S} \rightarrow \mathbb{D}}(0) = \varphi_{\mathbb{S} \rightarrow \mathbb{D}}\,, \\
      \phi_{\mathbb{R}^{(i)} \rightarrow \mathbb{D}}(0)) = \varphi_{\mathbb{R}^{(i)} \rightarrow \mathbb{D}}\,,
\label{app-eq: ode_initial}
\end{cases}
\end{equation}
where $f_0,\, f_1,\, \dots,\, f_{N_{\text{ref}}}$ are convolutional networks representing the motion evolution dynamics. Note when we perform one-shot motion transfer, no reference image is needed, and thus the above ODE system will reduce to a single ODE with an initial condition.

The integration of the above ODE system (Eqs.~(\ref{app-eq: ode_main}-\ref{app-eq: ode_initial})) results in the refinement process from the initial coarse motion deformations (Eq.~(\ref{app-eq: initial_dense})), to the fine-grained motion deformation estimation:
\begin{equation}
\begin{cases}
\Phi_{\mathbb{S} \rightarrow \mathbb{D}} = \phi_{\mathbb{S} \rightarrow \mathbb{D}}(1)\,, \\ 
\Phi_{\mathbb{R}^{(i)} \rightarrow \mathbb{D}} = \phi_{\mathbb{R}^{(i)} \rightarrow \mathbb{D}}(1)\,, \quad i = 1,\, \dots,\, N_{\text{ref}}\,.
\label{app-eq: final_dense}
\end{cases}
\end{equation}

Here, we prove the solution stated in Eq.~(\ref{app-eq: final_dense}) to the entire ODE system 1) exists, and 2) is unique.


\begin{proof}

For simplicity, let us first write Eqs.~(\ref{app-eq: ode_main}-\ref{app-eq: ode_initial}) in their matrix forms:
\begin{equation}
    \frac{d\,\Phi(\epsilon)}{d\, \epsilon} = F(\Phi(\epsilon),\, \epsilon), \quad  \Phi(0) = \Psi,
    \label{app-eq: ode_matrix}
\end{equation}
where
\begin{equation}
\begin{cases}
    \Phi = \big( \phi_{\mathbb{S} \rightarrow \mathbb{D}},\, \phi_{\mathbb{R}^{(1)} \rightarrow \mathbb{D}},\, \dots,\, \phi_{\mathbb{R}^{(N_{\text{ref}})} \rightarrow \mathbb{D}} 
    \big)^T \\
    F = \big( f_0,\, f_1,\, \dots,\, f_{N_{\text{ref}}} \big)^T
\end{cases}.
\end{equation}

~\\
\noindent \textbf{Existence.} 
Given the continuity nature of convolutional networks, for the non-linear function $F$, there $\exists$ functions $A(\epsilon)$, $U(\epsilon)$ which depend continuously on $\epsilon$ for $\epsilon \in [0,\, 1]$, such that the following linear system stands,
\begin{equation}
    \frac{d\,\Phi(\epsilon)}{d\, \epsilon} \leq 
    \frac{d\,\widetilde{\Phi}(\epsilon)}{d\, \epsilon} = A(\epsilon) \widetilde{\Phi}(\epsilon) + U(\epsilon), \quad  \widetilde{\Phi}(0) = \Psi.
    \label{app-eq: ode_matrix_linear}
\end{equation}
Since $A$, $U$ are continuous over $\epsilon \in [0,\, 1]$, they are therefore bounded and we could use the uniform norm:
\begin{equation}
    \Vert A \Vert := \underset{\epsilon \in [0,\, 1]}{\text{sup}}~ \vert A(\epsilon) \vert \,,\,\,~ \Vert U \Vert := \underset{\epsilon \in [0,\, 1]}{\text{sup}}~ \vert U(\epsilon) \vert \, ,
    \label{app-eq: upper_bound}
\end{equation}

First, we integrate both side of Eq.~(\ref{app-eq: ode_matrix_linear}) to find
 \begin{equation}
     \widetilde{\Phi}(\epsilon) = \Psi + \int_0^{\epsilon} \big[ A(s) \widetilde{\Phi}(s) + U(s) \big]~d s \,.
    \label{app-eq: ode_matrix_int}
 \end{equation}

By solving Eq.~(\ref{app-eq: ode_matrix_int}) by successive approximation with sufficiently large $N$ steps $k = 0,\, 1,\, 2, \, \dots ,\, N$, we obtain
\begin{equation}
     \widetilde{\Phi}_{k+1}(\epsilon) = \Psi + \int_0^{\epsilon} \big[ A(s) \widetilde{\Phi}_k(s) + U(s) \big]~d s \,,
    \label{app-eq: ode_matrix_disc}
\end{equation}
Since $\epsilon$ in this paper only denotes the refinement status and is not restricted to any physical unit meaning, we could further assume $\epsilon$ is in a smaller interval $[0 ,\, \beta]$, where $\beta < min( 1,\, 1/\Vert A \Vert) $. 
By subtracting the sequence in Eq.~(\ref{app-eq: ode_matrix_disc}), we have
 \begin{align}
     \widetilde{\Phi}_{k+1}(\epsilon) - \widetilde{\Phi}_{k}(\epsilon) & = \int_0^{\epsilon}  A(s) \big[\widetilde{\Phi}_{k}(s) - \widetilde{\Phi}_{k-1}(s) \big]~d s \nonumber \\
      & \leq \beta \Vert A \Vert   \int_0^{\epsilon} \big[ \widetilde{\Phi}_{k}(s) - \widetilde{\Phi}_{k-1}(s) \big] d s \nonumber \\
      & \leq \beta \Vert A \Vert \Vert \widetilde{\Phi}_{k} - \widetilde{\Phi}_{k-1} \Vert \nonumber \\
      & \leq c \Vert  \widetilde{\Phi}_{k} - \widetilde{\Phi}_{k-1} \Vert \,,
    \label{app-eq: ode_matrix_seq}
 \end{align}
where $c := \beta \Vert A \Vert < 1$. Note that since $\widetilde{\Phi}$ is continuous in $[0,\, \beta]$, $\int_0^{\epsilon} \big[ \widetilde{\Phi}_{k}(s) - \widetilde{\Phi}_{k-1}(s) \big] d s$ has the uniform norm, $\Vert \widetilde{\Phi}_{k} - \widetilde{\Phi}_{k-1} \Vert$, which is independent of $\epsilon$.

Because the right hand side of Eq.~(\ref{app-eq: ode_matrix_seq}) is independent of $\epsilon$, the sequence of the successive approximations is contracting
\begin{equation}
    \Vert \widetilde{\Phi}_{k+1} - \widetilde{\Phi}_{k} \Vert \leq c \Vert  \widetilde{\Phi}_{k} - \widetilde{\Phi}_{k-1} \Vert \,.
    \label{app-eq: inequal}
\end{equation}
Using the above inequality repeatedly we reach the following series
\begin{equation}
    \Vert \widetilde{\Phi}_{k+1} - \widetilde{\Phi}_{k} \Vert \leq c \Vert  \widetilde{\Phi}_{k} - \widetilde{\Phi}_{k-1} \Vert \leq \dots \leq c^k \Vert \widetilde{\Phi}_{1} - \widetilde{\Phi}_{0} \Vert \,.
    \label{app-eq: inequal_seq}
\end{equation}
Since $c \in (0,\, 1)$, the series $\sum \, c^k \Vert \widetilde{\Phi}_{1} - \widetilde{\Phi}_{0} \Vert $ converges. Therefore, by the Weierstrass M-test, the series $\sum \, \vert \widetilde{\Phi}_{k+1} - \widetilde{\Phi}_{k} \vert $ of continuous functions converges absolutely and uniformly in the interval $[0, \, \beta]$ to some continuous function. Meanwhile,
\begin{align}
\hspace*{-0.8cm}
    \sum\limits_0^N \big[ \widetilde{\Phi}_{k+1}(\epsilon) - \widetilde{\Phi}_{k}(\epsilon) \big] &= \big[ \widetilde{\Phi}_{N+1}(\epsilon) - \widetilde{\Phi}_{N}(\epsilon) \big] \nonumber \\ 
    & \quad + \big[ \widetilde{\Phi}_{N}(\epsilon) - \widetilde{\Phi}_{N-1}(\epsilon) \big]  \nonumber \\
    & \qquad + \dots + \big[ \widetilde{\Phi}_{1}(\epsilon) - \widetilde{\Phi}_{0}(\epsilon) \big] \nonumber \\
    & = \widetilde{\Phi}_{N+1}(\epsilon) - \widetilde{\Phi}_{0} \nonumber \\
    & = \widetilde{\Phi}_{N+1}(\epsilon) - \Psi \,.
    \label{app-eq: inequal_seq}
\end{align}
As a result, the sequence of continuous functions $\widetilde{\Phi}_N (\epsilon)$ converges uniformly in $[0, \, \beta]$ to some continuous function $\widetilde{\Phi}(\epsilon)$. Let the successive approximation steps to be approaching infite, i.e., $k \to \infty$ in Eq.~(\ref{app-eq: ode_matrix_disc}), we reach that $\widetilde{\Phi}(\epsilon)$ is the desired solution of the linear ODE system in Eq.~(\ref{app-eq: ode_matrix_linear}), which also indicates the the existence of $\Phi(\epsilon)$ in Eq.~(\ref{app-eq: ode_matrix}). In this last step, we used the uniform convergence to interchange limit and integral in Eq.~(\ref{app-eq: ode_matrix_disc}).


~\\
\noindent \textbf{Uniqueness.} 
Now we further demonstrate the uniqueness of the solution $\widetilde{\Phi}(\epsilon)$ to Eq.~(\ref{app-eq: ode_matrix_linear}). Same as in the above existence proof, we consider the interval $[0,\, \beta]$. 
Let us assume there exist two solutions to Eq.~(\ref{app-eq: ode_matrix_linear}), respectively $\widetilde{\Phi_a}(\epsilon)$ and $\widetilde{\Phi_b}(\epsilon)$. We now analyze their distance $W(\epsilon) := \widetilde{\Phi_a}(\epsilon) - \widetilde{\Phi_b}(\epsilon)$.

Deriving from Eq.~(\ref{app-eq: ode_matrix_linear}), we further have
\begin{equation}
    \frac{d\,W(\epsilon)}{d\, \epsilon} = A(\epsilon) W(\epsilon) \,, \quad W(0) \equiv 0 \,.
    \label{app-eq: solution_dist}
\end{equation}
Integrating both sides of the above equation, we obtain 
\begin{equation}
    W(\epsilon) = \int_0^{\epsilon} A(s) W(s) \, ds \,.
    \label{app-eq: solution_dist_int}
\end{equation}
Therefore, similar to the derivations in Eq.~(\ref{app-eq: ode_matrix_seq}), 
\begin{equation}
    W(\epsilon) \leq \beta \Vert A \Vert \Vert W \Vert \leq c \Vert W \Vert \,.
    \label{app-eq: ode_matrix_seq_ineq}
\end{equation}
Coupling with $c \leq 1$, Eq.~(\ref{app-eq: ode_matrix_seq_ineq}) indicates $\Vert W \Vert \equiv 0$, which means the two assumed solutions $\widetilde{\Phi_a}(\epsilon)$ and $\widetilde{\Phi_b}(\epsilon)$ are equal to each other for $ \forall \epsilon \in [0 ,\, \beta]$. This concludes the uniqueness of the solution of the ODE system in Eq.~(\ref{app-eq: ode_matrix}).

\end{proof}

\end{document}